\documentclass[sigconf, nonacm]{acmart}

\newcommand\vldbdoi{XX.XX/XXX.XX}

\newcommand\vldbpagestyle{plain} 

\usepackage{mathtools}      

\usepackage{xspace}
\newcommand{\name}{METER\xspace}

\usepackage{xcolor} 

\usepackage{colortbl}
\usepackage{amsopn}

\usepackage[linesnumbered,ruled,vlined]{algorithm2e}

\usepackage{makecell}
\usepackage{color}
\newcommand{\highlight}[1]{\textbf{#1}}

\newcommand{\blue}[1]{{\color{black}#1}}
\newcommand{\revise}[1]{{\color{black}#1}}

\newcommand{\powerpoint}[1]{\textbf{#1}}

\usepackage{enumitem}
\newcommand{\squishlist}
{
	\begin{list}{$\bullet$}
		{
			\setlength{\itemsep}{0pt}
			\setlength{\parsep}{3pt}
			\setlength{\topsep}{3pt}
			\setlength{\partopsep}{0pt}
			\setlength{\leftmargin}{1.5em}
			\setlength{\labelwidth}{1em}
			\setlength{\labelsep}{0.5em}
		}
	}
	
	\newcommand{\squishend}
	{
	\end{list}
}

\usepackage{multirow}
\usepackage{arydshln}
\usepackage{caption}
\usepackage{subcaption}
\usepackage{makecell}           
\usepackage{booktabs}  
\newsavebox{\measurebox}

\usepackage{multicol}

\newcommand{\unsim}{\mathord{\sim}}

\usepackage{url}

\usepackage{algorithmic}

\usepackage{dsfont}

\usepackage{balance}

\begin{document}


\title[\name: A Dynamic Concept Adaptation Framework for Online Anomaly Detection]{\name: A Dynamic Concept Adaptation Framework \\for Online Anomaly Detection}











\author{
{Jiaqi Zhu$^{1}$
, Shaofeng Cai$^2$, Fang Deng$^1$, Beng Chin Ooi$^2$, Wenqiao Zhang$^{3}$}\\
\fontsize{9}{9}\upshape\vspace{0.1em}
$^1$Beijing Institute of Technology  $^2$National University of Singapore   $^3$Zhejiang University\\
\fontsize{9}{9}\upshape\vspace{-0.4em}
\{jiaqi\_zhu, dengfang\}@bit.edu.cn, \{shaofeng, ooibc\}@comp.nus.edu.sg, wenqiaozhang@zju.edu.cn
}

\subtitle{[Technical Report]}


\begin{abstract}

\revise{
%
Real-time analytics and decision-making require online anomaly detection (OAD) to handle drifts in data streams efficiently and effectively.
Unfortunately, existing approaches are often constrained by their limited detection capacity and slow adaptation to evolving data streams, inhibiting their efficacy and efficiency in handling \textit{concept drift}, which is a major challenge in evolving data streams.
In this paper, we introduce \name, a novel dynamic concept adaptation framework that introduces a new paradigm for OAD.
\name addresses concept drift by first training a base detection model on historical data to capture recurring \textit{central concepts}, and then learning to dynamically adapt to \textit{new concepts} in data streams upon detecting concept drift.
Particularly, \name employs a novel \textit{dynamic concept adaptation} technique that leverages a hypernetwork to dynamically generate the parameter shift of the base detection model, providing a more effective and efficient solution than conventional retraining or fine-tuning approaches.
Further, \name incorporates a lightweight drift detection controller, underpinned by evidential deep learning, to support robust and interpretable concept drift detection.
We conduct an extensive experimental evaluation, and the results show that \name significantly outperforms existing OAD approaches in various application scenarios.
}

\end{abstract}

\maketitle

\pagestyle{\vldbpagestyle}



\section{Introduction}\label{sec:introduction}
Anomaly detection (AD), the process of identifying data samples that significantly deviate from the majority, 
plays a critical role in various systems by facilitating a deeper understanding of data, uncovering hidden anomalies, and enabling the implementation of appropriate measures to address associated concerns~\cite{chandola2009anomaly,li2022unsupervised,toliopoulos2020proud,zenati2018adversarially,schmidl2022anomaly,boniol2021unsupervised}.
While many methods have been developed for detecting anomalies in static data~\cite{zong2018deep,kim2020rapp,lai2019robust,yoon2022adaptive,mirsky2018kitsune}, the challenge of identifying anomalies in evolving data streaming applications has not been adequately addressed.

In real-world scenarios, data is often subject to constant updates and changes, primarily due to the dynamic nature of data sources (e.g., stock markets, traffic flow, social media), underlying infrastructures (e.g., IoT devices, cloud-based equipment, edge servers),  or the influence of many other factors~\cite{pang2021deep,ruff2021unifying,chawathe1997meaningful}.
Such dynamic nature of data requires real-time processing and analytics to effectively manage and respond to these changes.
In particular, online processing and analysis of anomalies in evolving data streams have become increasingly essential for maintaining data quality and security across various domains.
Notably, online anomaly detection (OAD) on evolving data streams enables real-time and better-informed decision-making, thereby ensuring data integrity and security while supporting the daily operations across business sectors~\cite{kloft2010online,savage2014anomaly,kloft2012security,paparrizos2022tsb}.


\begin{figure*}[t]
    \centering
    \includegraphics[width=0.88\linewidth]{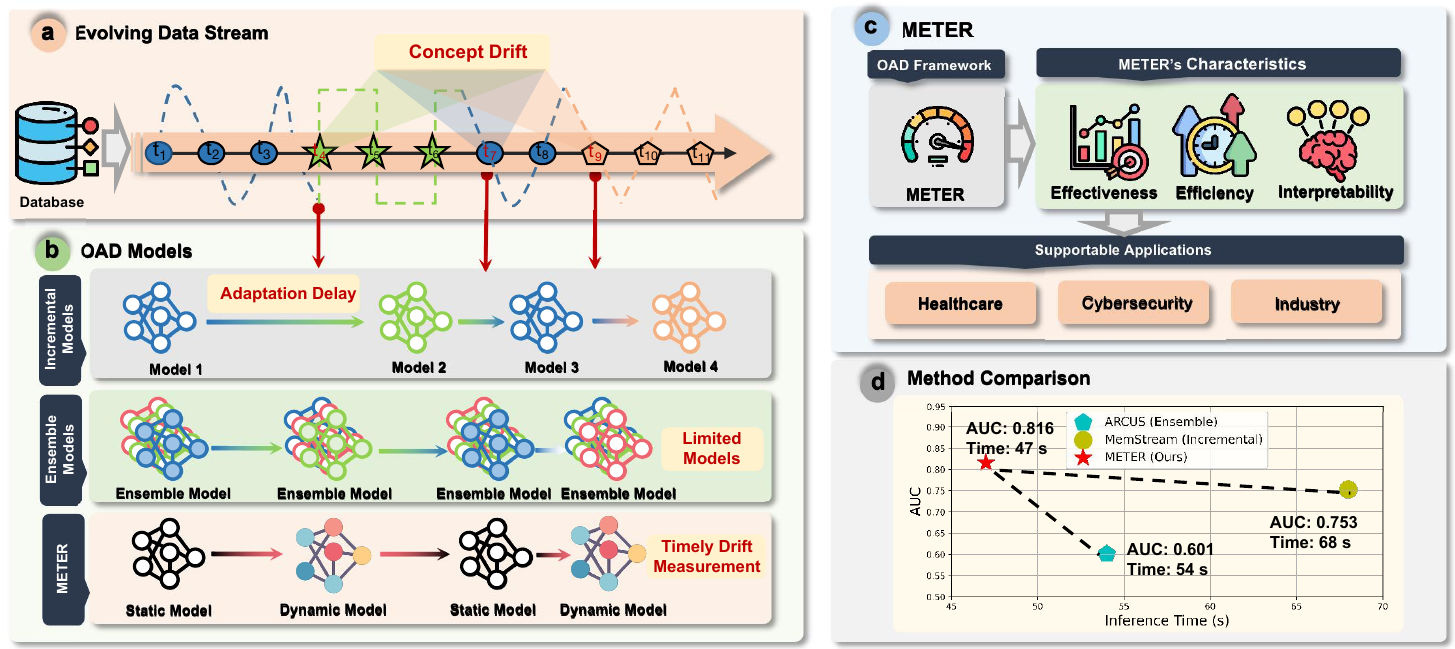} \vspace{-1mm}
    \caption{\blue{(a) The illustration of the evolving data stream with concept drifts in online anomaly detection (OAD).
    (b) The three main types of OAD approaches.
    (c) The functionalities of \name supporting various applications.
    (d) The comparison of our \name to state-of-the-art OAD approaches on the real-world dataset INSECTS~\cite{souza2020challenges}, where Memstream~\cite{bhatia2022memstream}  and ARCUS~\cite{yoon2022adaptive} are representative incremental and ensemble approaches: \texttt{AUC: 0.816 (\name) > 0.753 (Memstream) > 0.601 (ARCUS)}, Inference Time: 47 s (\name) < 54 s(Memstream) < 68 s (ARCUS).}}
    \label{fig:intro}
    \vspace{-3mm}
\end{figure*}

Evolving data streams are characterized by high-dimensional and heterogeneous data arriving continuously.
Such ever-changing data streams make models trained on historical data outdated quickly.
This phenomenon, commonly referred to as \textit{concept drift}, as illustrated in Figure~\ref{fig:intro} (a), presents great challenges for the conventional AD approaches ~\cite{deng2022anomaly,gong2019memorizing,kim2020rapp,lai2019robust}, which train their detection models once on static data and assume the models keep functioning after deployment.
This is not practical as they are built on the unrealistic assumption that all normal patterns are known prior to the model deployment.
\revise{
For instance, in the transaction monitoring of financial institutions, normal transaction patterns and behaviors will gradually change over time due to seasonal variations, shifts in market trends, new trading strategies, etc.
Suppose the transaction records of a financial institution are stored and managed in a database system, detection models of conventional static AD approaches are typically trained by retrieving historical transaction data, which can perform well during a specific time period.
However, once \textit{concept drift} occurs, these static detection models could no longer effectively detect new abnormal behaviors.
Therefore, AD approaches that can respond timely to drifts in data distributions become imperative.
}


Recently, several preliminary attempts have been made to support OAD in evolving data streams~\cite{bhatia2022memstream, bhatia2021mstream,yoon2021multiple,yoon2020ultrafast, guha2016robust,yoon2022adaptive,mirsky2018kitsune,kieu2019outlier,boniol2021sand,lu2022matrix,tran2020real}.
One popular approach for handling streaming data is based on incremental learning~\cite{bhatia2022memstream, bhatia2021mstream,yoon2021multiple,yoon2020ultrafast, guha2016robust,boniol2021sand}.
These methods typically construct an initial model, which is then incrementally updated as new data arrives.
However, as shown in Figure~\ref{fig:intro} (b), incremental approaches often require a considerable amount of time to adapt to new \textit{concepts}~\cite{yoon2022adaptive} and do not provide performance guarantees for handling arbitrary concept drifts.
\revise{Moreover, incremental learning methods typically suffer from inefficiency issues, as they need to update models to accommodate the changing concepts, either by retraining, fine-tuning models, or updating certain model statuses~\cite{bhatia2022memstream,yoon2021multiple,yoon2020ultrafast}.}
Another line of research in OAD focuses on the ensemble-based approaches~\cite{yoon2022adaptive,mirsky2018kitsune,kieu2019outlier}, which involve training a combination of models to account for respective changing concepts.
However, as shown in Figure~\ref{fig:intro} (b), the detection capacity of these model ensembles is constrained by the number of pre-trained models, and the model ensemble needs to adapt to drifted concepts frequently as new data arrives.
Additionally, ensemble-based approaches require substantially more computation resources to maintain multiple models, further exacerbating the inefficiency issue.
These limitations hinder existing approaches from meeting the requirements of supporting OAD in evolving data streams effectively and efficiently.

In particular, the deployment of an OAD framework in real-world applications necessitates not only detection effectiveness, but also efficiency and interpretability~\cite{bhatia2021mstream,yoon2021multiple,yoon2022adaptive}.
In terms of efficiency, the detection framework must provide accurate and timely decisions, and meanwhile, maintain a time complexity that does not increase linearly with the growing volume of data.
\revise{
As for interpretability, the ability of the detection framework to provide reliable uncertainty estimates for the detection results is necessary for users in critical applications such as healthcare and finance, which improves human understanding and engenders user trust in the framework~\cite{cai2021arm,chen2022adaptive,zheng2020tracer}.
}
Therefore, an ideal OAD framework should address all three crucial criteria, delivering effective, efficient and more interpretable detection in evolving data streams.

In this paper, we introduce a novel dyna\textbf{\underline{M}}ic conc\textbf{\underline{E}}pt adap\textbf{\underline{T}}ation fram\textbf{\underline{E}}wo\textbf{\underline{R}}k (\name) for online anomaly detection in evolving data streams.
Different from the conventional incremental learning and ensemble-based approaches, \name presents a new OAD paradigm that effectively, efficiently, and more interpretably addresses the concept drift challenge inherent in evolving data streams.
\name is built upon the key observation that static historical data typically encompass the majority of anomaly patterns and thus comprises the recurring \textit{central concepts}, whereas new arriving data streams, although may deviate from the main patterns occasionally, usually only drift slightly away from central concepts.
\blue{Taking into account this observation in the framework design, \name first trains a base detection model to capture central concepts, and then, learns to adapt to \textit{new concepts} once detecting concept drift without further training, fine-tuning, or updating model statuses as required in conventional approaches~\cite{mirsky2018kitsune,kieu2019outlier,bhatia2022memstream,yoon2021multiple,yoon2020ultrafast}.
}

\blue{To improve efficiency and effectiveness, we further introduce a novel lightweight controller to detect whether concept drift occurs in the current data stream on a per-input basis, and design a novel \textit{dynamic concept adaptation} technique that dynamically generates the corresponding \textit{parameter shift} of the base detection model for the current input via a hypernetwork ~\cite{ha2016hypernetworks}.
After training, \name can dynamically generate the \textit{parameter shift} for the current data stream to handle the concept drifts on the fly and enhance the predictive performance of the base detection model.
To support interpretability,} this controller is derived from the \textit{evidential deep learning} (EDL) theory~\cite{ng2011dirichlet,sensoy2018evidential}, which enables efficient and high-quality uncertainty modeling for the detection of concept drift, thereby supporting interpretable anomaly detection.
\blue{Meanwhile, dynamic concept adaptation via the hypernetwork is more effective in handling concept drift, as the detection model can adapt to new concepts in an input-aware and timely manner, and is more efficient, since this novel approach requires no frequent model retraining or fine-tuning for the concept adaptation as in approaches based on incremental or ensemble techniques~\cite{mirsky2018kitsune,kieu2019outlier,bhatia2022memstream,yoon2021multiple,yoon2020ultrafast}.}


Our \name comprises the following main components:
\textbf{(i)} Static Concept-aware Detector (SCD), an unsupervised deep autoencoder (AE)~\cite{kingma2013auto} to minimize the reconstruction error via compressed representations.
SCD is pretrained on historical data to model the central concepts.
\textbf{(ii)} Intelligent Evolution Controller (IEC), a concept detection controller that dynamically models \textit{concept uncertainty} via \textit{evidential deep learning}~\cite{sensoy2018evidential}. IEC detects concept drift for the current data stream.
\textbf{(iii)} Dynamic Shift-aware Detector (DSD), 
a hypernetwork that dynamically updates SCD in an input-aware manner with the parameter shift upon the detection by concept drift by the IEC.
DSD streamlines dynamic concept adaptation, as the concept drift can now be handled efficiently and more effectively by learning only the parameter shift, in a way reminiscent of residual learning~\cite{wu2019wider,he2016deep}.
\textbf{(iv)} Offline Updating Strategy (OUS), a strategy introduced to enhance SCD with new central concepts.
\blue{To improve efficiency, OUS only updates \name with new concepts when the recent data streams contain markedly different concepts from the existing central concepts.}
To achieve this, OUS employs a sliding window to aggregate the statistics of concept uncertainty to determine whether an update is needed.
With these modules, \name delivers a more effective, efficient and interpretable OAD framework.
We summarize our main contributions as follows:

\begin{itemize}[leftmargin=*]
    \item
    We propose a novel unsupervised OAD framework \name for evolving data streams, 
    which offers a new OAD paradigm that effectively, efficiently, and interpretably addresses the concept drift challenge inherent in evolving data streams.

    \item
    We incorporate into our \name 
    a lightweight Intelligent Evolution Controller (IEC) for detecting concept drift in evolving data streams on a per-input basis, which enables efficient and high-quality uncertainty modeling for the detection of concept drift, thereby supporting interpretable anomaly detection.
    

    \item 
    We develop a Dynamic Shift-aware Detector (DSD) that dynamically updates the base detection model with the parameter shift via a hypernetwork in an input-aware manner.
    DSD streamlines dynamic concept adaptation and handles concept drift efficiently and more effectively.



    \item Through extensive experiments, we demonstrate that our \name framework efficiently detects various types of anomalies with high accuracy and interpretability,  and meanwhile, outperforms existing incremental and ensemble learning methods for online anomaly detection.

\end{itemize}

The remainder of this paper is organized as follows: Section~\ref{sec:preliminary} introduces the preliminaries, including key concepts and background information.
Section~\ref{sec:methodology} presents our Detective framework, providing a comprehensive overview of its modules, optimization schemes, and discussions on its effectiveness, efficiency, and \blue{interpretability}.
In Section~\ref{sec:experiment}, we showcase experimental results and evaluate the framework's effectiveness and \blue{interpretability}.
Section~\ref{sec:related work} reviews related work in the field, and finally, Section~\ref{sec:conclusion} concludes the paper.

\section{Preliminaries}
\label{sec:preliminary}

In this section, we outline the fundamental concepts and techniques related to anomaly detection, concept drift, hypernetwork, and evidential deep learning.
We begin by formally defining the problem of anomaly detection and concept drift, followed by a detailed description of the two key techniques that form the basis of our \name framework, namely hypernetwork and evidential deep learning.
Scalars, vectors, and matrices are denoted as $x$, $\vec{x}$ and $X$ respectively.

\noindent
\highlight{Anomaly Detection and Concept Drift.}
\blue{The objective of \textit{Anomaly detection} is to identify data samples that deviate from the majority or exhibit unusual patterns.
In particular, the focus of this paper is on \textit{online anomaly detection} (OAD), which is to detect anomalies in data streams in real-time, as formulated in Definition~\ref{def:OAD} below.
Further, our framework focuses on unsupervised anomaly detection in data streams, where no labeling information is available.}

  
\blue{In real-world scenarios, data is often dynamic and subject to constant changes~\cite{lu2018learning}, a phenomenon known as \emph{concept drift}, where the statistical or distributional properties of the data within a certain domain change over time, as formally defined in Definition~\ref{def:concept drift}. }
This change in distribution can lead to a significant drift in the patterns and relationships of the data, presenting challenges in detecting anomalous events. 
\blue{
\begin{definition}[Online anomaly detection]
    \label{def:OAD}
    Considering an incoming data stream $\mathcal{X} =\left\{ \vec{x}_1, \dots, \vec{x}_t, \dots \right \} $, where each entry $ \vec{x}_t=\left( x_{t1},\dots,x_{td} \right) $ comprises $d$ attribute fields that can be either categorical or numerical features, \textit{online anomaly detection} (OAD) aims to predict whether a data sample $\vec{x}_t$ in the incoming data stream is anomalous or not at each time step $t$.\vspace{-2mm}
\end{definition}}
\blue{
\begin{definition}[Concept drift]
    \label{def:concept drift}
    A concept drift occurs at time step $t$ if the underlying joint probability $P\left( \vec{x}, y \right)$ of input data $\vec{x}$ and the corresponding label $y$ changes at time $t$, that is, $P \left( \vec{x}_t,y_t \right) \ne P\left( \vec{x}, y \right)$.
\end{definition}
}

To address this challenge, the model needs to capture the dynamic behavior of the data stream and effectively adapt to drifted concepts over time.
Formally, OAD can be achieved by dynamically computing an anomaly score $\mathcal{M}( \vec{x}_t; \Theta_m) $ for each time step $t$, using a detection model $\mathcal{M}(\cdot)$ parameterized by $\Theta_m$.


\vspace{1mm}
\noindent
\highlight{Hypernetwork.}
A \textit{hypernetwork} is a neural network that generates the weights for another neural network, known as the \textit{primary network}.
Hypernetworks are typically used to condition the parameters of the primary network on certain input features, allowing the primary network to adapt its behavior based on the input data.
In our \name framework, we leverage a hypernetwork to learn the parameter shift for the Static Concept-aware Detector (SCD), enabling it to adapt to the evolving data stream with new concepts in an instance-aware manner.

The basic architecture of SCD is a multi-layer perception (MLP) based autoencoder~\cite{kingma2013auto}.
To measure the parameter shift of this SCD MLP with $N_l$ layers, the hypernetwork models each MLP layer as a matrix $K^{(n)} \in \mathbb{R}^{N_{in}\times N_{out}}$, where $N_{in}$ and $N_{out}$ are the number of input and output neurons of the $n$-th layer of the MLP respectively.
The generation process of matrix $K^{(n)}$ can then be regarded as matrix factorization as below:
 \begin{align}
    K^{(n)} = \xi(\vec{r}^{(n)};\Theta_h), \forall n=  1, \cdots, N_l\,.
\end{align}
where $\vec{r}^{(n)}$ is a vector, $\xi(\cdot)$ is a randomly initialized MLP, and $\Theta_h$ is the parameters of $\xi(\cdot)$.
Such a generation process enables the gradients to backpropagate to $\vec{r}^{(n)}$ and $\xi(\cdot)$ for effective end-to-end training.
In this way, the parameter shift of the SCD can be adaptively measured by $\vec{r}^{(n)}$ and $\xi(\cdot)$ instead of $K^{(n)}$ directly.

\vspace{1mm}
\noindent
\highlight{Evidential Deep Learning.}
\textit{Evidential Deep Learning} (EDL) ~\cite{sensoy2018evidential} is a probabilistic deep learning approach that interprets the categorical predictions of a neural network as a \textit{distribution} over class probabilities by placing a Dirichlet prior upon the class probabilities.
This allows the network to provide not only point estimates for the detection but also \textit{uncertainty estimates} for each prediction. 
In \name, we utilize EDL to measure concept uncertainty, enabling the Intelligent Evolution Controller (IEC) to dynamically evolve the Static Concept-aware Detector (SCD) to the Dynamic Shift-aware Detector (DSD) based on the concept uncertainty of the input data.


Considering a general C-class classification task, given an instance $\vec{x}$, a standard DNN with the softmax operator is usually adopted after processing features of $\vec{x}$ to convert the predicted logit vector into the class probability vector $\vec{p}$.
When using EDL for such a DNN, a Dirichlet distribution 
is placed over the categorical likelihood $\vec{p}$ to model the probability density of each possible $\vec{p}$.
The probability density function of $\vec{p}$ for $\vec{x}$ is obtained by:
\begin{equation}
\begin{aligned}
\!\!\!\! \!\!P(\vec{p}|\vec{x};\Theta_e)=Dir(\vec{p}|\vec{\alpha})=
 \begin{cases} 
\frac{\Gamma (\sum_{c=1}^C \alpha_{c})}{\prod_{c=1}^C\Gamma(\alpha_{c})}\prod_{c=1}^C p_{c}^{\alpha_{c}-1}, \!\! \!\! & \mbox{if } \vec{p} \in \Delta^C\\
\quad\quad\quad\quad0 \quad\quad\quad\quad\quad, &  \!\!\mbox{otherwise}
\end{cases} 
 \label{edl}
\end{aligned}
\end{equation}

\noindent
where $\vec{\alpha}$ is the parameters of the Dirichlet distribution $Dir(\vec{p}|\vec{\alpha})$ for the sample $\vec{x}$, $\Gamma(\cdot)$ is the Gamma function, and $\Delta^C$ is the $C$-dimensional unit simplex: $\Delta^C$=$\{\sum_{c=1}^C p_{c}=1 \ {\rm and} \ 0\leq p_c \leq 1 $\}.
Particularly, $\vec{\alpha}$ can be modeled as $\vec{\alpha}=g(f(\vec{x},\Theta_f))$, 
where $f(\cdot)$ is another DNN model, and $g(\cdot)$ is the exponential function to keep $\vec{\alpha}$ positive.
In this way, the prediction of the sample $\vec{x}$ can be interpreted as a distribution over the probability, \emph{i.e.}, the concept uncertainty modeled using IEC, rather than the simple and unreliable predictive uncertainty~\cite{sensoy2018evidential,ng2011dirichlet}.

\section{Methodology}
\label{sec:methodology}
This section first presents the overview of \name, which is designed to adaptively detect the online anomaly data under concept drift. We then elaborate on each module and introduce the optimization scheme. We further discuss the effectiveness, efficiency and interpretability of our \name.  

\begin{figure*}[t]
    \centering 
    \includegraphics[width=0.98\linewidth]{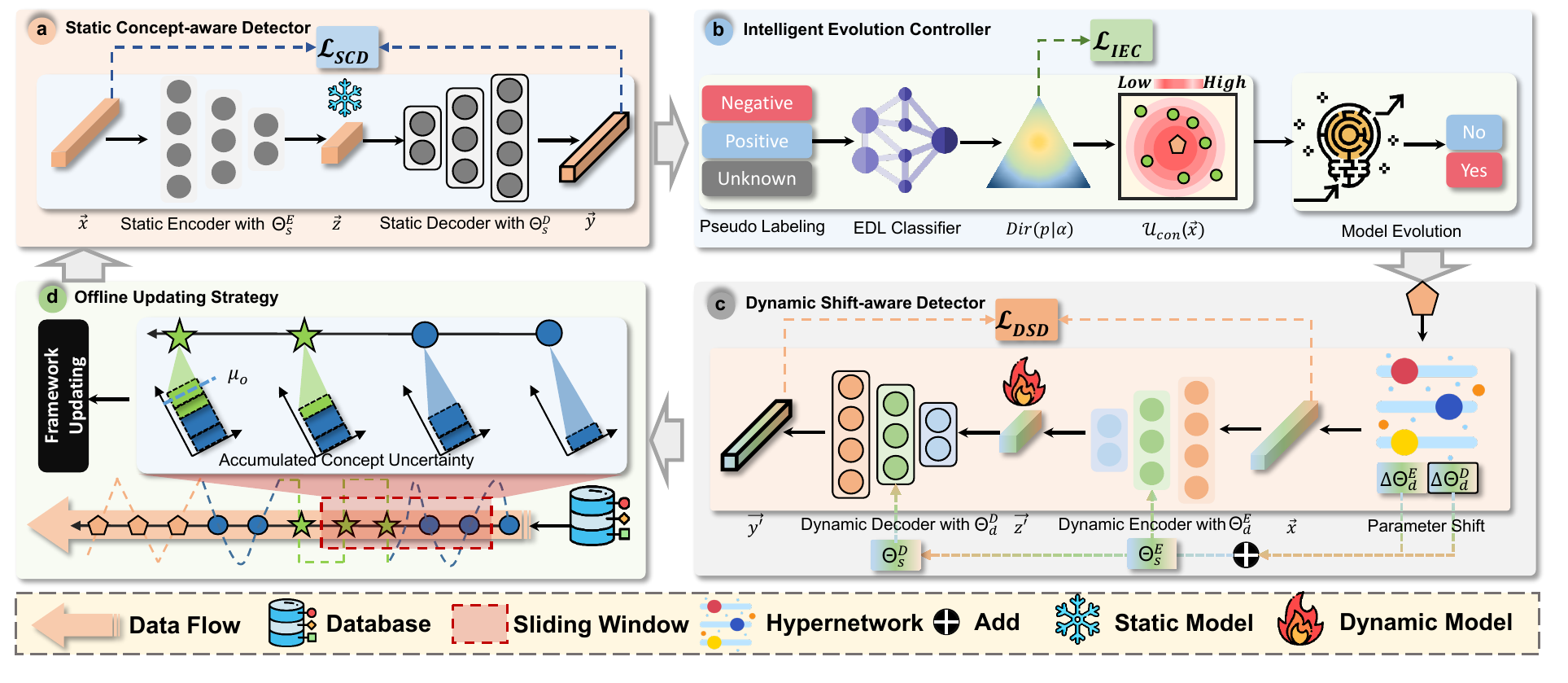}  \vspace{-3mm}
    \caption{Overview of the proposed \name. (a) Static Concept-aware Detector (SCD) is first trained on historical data to model the central concepts. (b) Intelligent Evolution Controller (IEC) timely measures the concept uncertainty to determine the necessity of dynamic model evolution. (c) Dynamic Shift-aware Detector (DSD) dynamically updates SCD with the instance-aware parameter shift by considering the concept drift.  (d) Offline Updating Strategy (OUS) introduces an effective framework updating strategy according to the accumulated concept uncertainty in a given sliding window. }
    \label{fig:framework}
\end{figure*}

\subsection{Overview}\label{sec:overview}
The overview of \name is illustrated in Figure~\ref{fig:framework}, which depicts the entire proposed pipeline. 
The main intuition is that the evolving data stream with different concepts should be identified and measured for dynamic model evolution.
To achieve this goal, we propose four modules, namely Static Concept-aware Detector (Sec.~\ref{sec:scd}), Intelligent Evolution Controller (Sec.~\ref{sec:isc}), Dynamic Shift-aware Detector (Sec.~\ref{sec:dsd}) and Offline Updating Strategy (Sec.~\ref{sec:ous}).
We will elaborate on these modules in the following subsections.

\subsection{Architecture}\label{sec:architecture}

\subsubsection{Static Concept-aware Detector}
\label{sec:scd}
The Static Concept-aware Detector (SCD) aims to detect the anomaly data from the static data stream with the central concepts, \emph{i.e.}, measuring the overall distribution of the historical data stream.
In anomaly detection tasks, 
the superior performance and unsupervised learning nature of autoencoders make them widely applicable involving unlabeled data~\cite{ audibert2020usad,abdulaal2021practical}.
Thus, we employ an autoencoder as the static base model that is trained on the historical data stream with central concepts to detect the central types of anomalies.


\noindent\textbf{Static Autoencoder.} 
An autoencoder is a deep neural network that learns to reconstruct its inputs.
Concretely, given the input instance $\vec{x}$ in the data stream $\mathcal{X}$, a static autoencoder with parameters $\Theta_s=(\Theta_s^E,\Theta_s^D)$ learns to reconstruct $\vec{x}$ by:
 \begin{align}
    \vec{x} \Rightarrow \mathcal{E}_s(\vec{x};\Theta_s^E)=\vec{z} \Rightarrow  \mathcal{D}_s(\vec{z};\Theta_s^D)=\vec{y}, \quad s.t. \quad \vec{x} \thickapprox \vec{y}
\end{align}
\noindent
where the static autoencoder comprises two main components, an encoder $\mathcal{E}_s(\cdot)$ and a decoder $\mathcal{D}_s(\cdot)$ with parameters $\Theta_s^E$ and $\Theta_s^D$, respectively.
The encoder $\mathcal{E}_s(\cdot)$ compresses the representation of the input $\vec{x}$ into a latent representation vector $\vec{z}$, and then the decoder $\mathcal{D}_s(\cdot)$ reconstructs the original input into $\vec{y}$ using $\vec{z}$.

Essentially, an autoencoder attempts to learn the identity function of the original data distribution.
Therefore, certain constraints are placed on the neural network, forcing it to learn meaningful concepts and relationships among features of $\vec{x}$.
As such, the static autoencoder gains the capability to reconstruct unseen inputs during the reconstruction training from the same data distribution $\mathcal{X}$.
For anomaly detection, in particular, if the input does not belong to the central concepts learned from $\mathcal{X}$, then we expect the reconstruction to have a larger error.

\subsubsection{Intelligent Evolution Controller.}
\label{sec:isc}
While the small/large reconstruction error of the static autoencoder indicates whether the input data is normal/abnormal, this result becomes unreliable when the data is out-of-distribution~\cite{haroush2021statistical}, especially in the presence of concept drift.
To determine whether the concept of a given input $\vec{x}$ belongs to central concepts from historical data or new concepts from evolving data streams,  we introduce an Intelligent Evolution Controller (IEC).
IEC can adaptively and timely examine the necessity of evolving the static detector by detecting the concept drift in the current data stream.

\noindent\textbf{Pseudo Labeling Strategy.}
IEC is a lightweight evidential classifier with parameters $\Theta_c$ trained with the high-confidence pseudo labels from the static autoencoder.
The rationale behind introducing pseudo labels in \name is to treat the input instance
that the static autoencoder cannot reconstruct well as negative with label 1, while treating the well-reconstructed input as positive with label 0, so that IEC can learn to detect whether the current input belongs to the central concepts already captured in SCD or new concepts:
\begin{equation}
\begin{aligned}
\!\!\!\!\!\! \tilde{y}(\vec{x})=
 \begin{cases} 
1 (\texttt{Positive}), \!\! & \mbox{if } L_{2}(\vec{x},\vec{y};\Theta_s) >\mu_p \mbox{ and } \mathcal{U}_{ctr}(\vec{x})\leqslant\mu_{e}\\
0 (\texttt{Negative}), \!\! & \mbox{if } L_{2}(\vec{x},\vec{y};\Theta_s) \leqslant \mu_p \mbox{ and } \mathcal{U}_{ctr}(\vec{x})\leqslant \mu_{e}\\
\textbf{-}(\texttt{Unknown}), & { \mathcal{U}_{ctr}(\vec{x} ) > \mu_{e}}
\end{cases} 
\label{pseudo label}
\end{aligned}
\end{equation}
\noindent where $L_{2}(\vec{x},\vec{y};\Theta_s)$ is the reconstruction error~\cite{sabokrou2016video}  of the input instance given the static autoencoder computed by taking the root mean squared error between $\vec{x}$ and the reconstructed output $\vec{y}$; 
$\mu_p$ is a predefined pseudo labeling threshold determined by setting a proportion of the sorted reconstruction error over all training samples;
$\tilde{y}(\vec{x})$ denotes the pseudo label of $\vec{x}$;
$\mathcal{U}_{ctr}(\vec{x})$ is the concept uncertainty that determines the necessity of model evolution, which will be introduced in detail next;
$\mu_e$ is a predefined threshold to ensure that the static autoencoder is only trained with high-confidence samples with a low concept uncertainty $\mathcal{U}_{ctr}(\vec{x})$, \emph{i.e.}, $\vec{x}$ is not involved in training if the $\mathcal{U}_{ctr}(\vec{x}) > \mu_{e}$ ($\tilde{y}(\vec{x})$=\texttt{Unknown}).

\noindent\textbf{Concept Uncertainty Estimation.} We first give the definition of the predicted probability of the evidential learning-based model. Considering the instance $\vec{x}$, the predicted probability  $\hat{P}(y=c|\vec{x};\Theta_c)$ of the class $c$ (0 or 1) following Eq.~(\ref{edl}) by marginalizing over $\vec{p}$ is:
\begin{equation}
\begin{aligned}
&\hat{P}(y=c|\vec{x};\Theta_c)= \int P(y=c|\vec{p};\Theta_c)P(\vec{p}|\vec{x};\Theta_c)d\vec{p}\\
&=\frac{\alpha_c}{\sum_{k=1}^C\alpha_k}=\frac{g(f_c(\vec{x}))}{\sum_{k=1}^C g(f_k(\vec{x}))}=\mathbb{E}[Dir(\vec{p}_c|\vec{\alpha})]\,,
\label{prediction}
\end{aligned}
\end{equation}
\noindent
where $g(\cdot)$ adopts the exponential function so that the softmax-based prediction can be interpreted as the expectation of the Dirichlet distribution.

For the evidential IEC trained using the supervision of pseudo labels from central concepts, when the concepts of current samples are obviously distinct from central concepts, \emph{i.e.}, the data is out-of-distribution, the evidence collected for these samples will be insufficient, as the controller now lacks the knowledge of such new concepts~\cite{sensoy2018evidential,ng2011dirichlet}.
Built upon this, we introduce the uncertainty resulting from the lack of evidence, called \textit{concept uncertainty}, to measure the extent to which the current concept drifts.
Formally, concept uncertainty $\mathcal{U}_{ctr}$ of the instance $\vec{x}$ is defined as:
\begin{equation}
\begin{aligned} 
&\mathcal{U}_{ctr}(\vec{x};\Theta_c) = \sum_{c=1}^C \hat{P}(y=c|\vec{x};\Theta_c) \left(\Phi(\alpha_{c}+1\right)- \\& \Phi(\sum_{k=1}^C \alpha_{k}+1))-\sum_{c=1}^C \hat{P}(y=c|\vec{x};\Theta_c) {\rm log} \hat{P}(y=c|\vec{x})\,,
 \label{domain}
\end{aligned}
\end{equation}
where $\Phi(\cdot)$ is the digamma function. Here, we use mutual information to measure the spread of Dirichlet distribution on
the simplex following~\cite{sensoy2018evidential}. A higher $\mathcal{U}_{ctr}(\vec{x})$ indicates a larger concept uncertainty, \emph{i.e.}, the Dirichlet distribution is widely dispersed across the probability simplex.

When deriving a higher concept uncertainty $\mathcal{U}_{ctr}(\vec{x}) > \mu_e$, the OAD prediction from the static autoencoder is interpreted as unreliable.
In this way, IEC measures instance-aware concept shift and enables the dynamic shift-aware detector to further process the current input instance.

\subsubsection{Dynamic Shift-aware Detector}
\label{sec:dsd}
As discussed above, the SCD module can effectively measure the central concepts derived from the historical data stream.
However, SCD may fail to support accurate detection when the statistical or distributional properties of the data stream shift in new data streams over time.
In this sense, modeling new concepts 
when concept drift occurs is critical, which enhances the detector for accurate detection across time.
To handle evolving data streams, we introduce a Dynamic Shift-aware Detector (DSD) that effectively enhances the static autoencoder with the new concepts in the current data stream.

\noindent\textbf{Parameter Shift Measurement.}
To measure concept drift, we leverage a hypernetwork to learn the parameters shift $\Delta\Theta_d$ for SCD with the same set of parameters, namely the encoder parameter shift $\Delta\Theta_d^E$ and the decoder parameter shift $\Delta\Theta_d^D$.
However, as shown in Eq.(1), the original hypernetwork only utilizes a randomly initialized $\vec{r}^{(n)}$ to generate the model parameters, which \blue{lacks} the interaction between the parameter generation process and the current input instance.
To measure the instance-aware parameter shift, we propose to model the parameter shift by replacing the $\vec{r}^{(n)}$ with representations of the current input instance.
Specifically, given the input $\vec{x}$, the hypernetwork first employs a layer-specific subnetwork $E^{(n)}(\cdot)$ that captures $\vec{x}$ related features as ${\vec{e}}^{(n)}$ for the parameter shift generation of the $n$-th layer parameters.
To reuse features and reduce model parameters, different layers share one encoder network $E_{\rm{share}}(\cdot)$ while employing different linear layers to get the layer-specific representation vector ${e}^{(n)}$ in the hypernetwork:
\begin{align}
\label{eq:lightweight_encoder}
    {\vec{e}}^{(n)} = E^{(n)}(\vec{x}) = {L}^{(n)}_{\rm{layer}}({E}_{\rm{share}} (\vec{x})), \rm{n }=  1, \cdots, \mathcal{N}_d,
\end{align}
\noindent
where $N_d$ is the number of the encoder and decoder layers,
and ${L}^{(n)}_{\rm{layer}}(\cdot)$ is the linear layer to transform the output of ${E}_{\rm{share}}(\cdot)$ to the $n$-th layer features. 

The parameter shift of the $n$-th layer of the SCD encoder/decoder can be formatted as a matrix $K^{(n)} \in \mathbb{R}^{N_{in}\times N_{out}}$, where $N_{in}$ and $N_{out}$ are the number of input neurons and output neurons respectively. 
Then, we transform the input-aware feature vector ${\vec{e}}^{(n)}$ into the parameter shift of the corresponding layer.
Specifically, the hypernetwork further employs the following two MLP layers to generate the parameter shift of the $n$-th layer:
\begin{align}
\label{eq:kernal_generation_detail}
\vspace{-0.1cm}
\begin{gathered}
    {W}^{(n)} = (W_1{\vec{e}}^{(n)} + \vec{b}_1)W_2 + \vec{b}_2, \\
    K^{(n)} = {W}^{(n)} + {{\vec{b}}}^{(n)},
\end{gathered}
\vspace{-0.1cm}
\end{align}
\noindent
where $W_1$ and $W_2$ are weights of the two MLP layers of the hypernetwork respectively, $\vec{b}_1$, $\vec{b}_2$ and  $\vec{b}$ are the biases.
Altogether, parameter shifts of the encoder and decoder for the static autoencoder can be denoted as $\Delta\Theta_d^E$ and $\Delta\Theta_d^D$.

\begin{algorithm}[t]
\small
\renewcommand{\algorithmicrequire}{\textbf{Input:}}  
\renewcommand{\algorithmicrequire}{\textbf{Initialization:}}

\begin{flushleft}
  \caption{\name Training}
    \textbf{Input}: 
    Small subset of stream data $\mathcal{X}_s$ \\
    \textbf{Output}: SCD, IEC, DSD parametrized by $\Theta_s$, $\Theta_c$  and $\Theta_d$\\
    \textbf{Initialization}: Randomly initialized $\Theta_s$, $\Theta_c$  and $\Theta_d$\\
    
     \Repeat {Convergence}{Randomly sample a minibatch\\
        \If{SCD not trained in \name}{Update parameters $\Theta_s$ via minimizing $\mathcal{L}_{SCD}$ using Eq.(12)} 
        \Else {
        Pseudo-labeling samples using Eq.(4)
        }
        }        
  \Return{SCD with parameters $\Theta_s$}.

   \Repeat {Convergence}{Randomly sample a minibatch\\
    Update parameters $\Theta_c$ via minimizing $\mathcal{L}_{IEC}$ using Eq.(13)
    
    Update parameters $\Theta_d$ via minimizing $\mathcal{L}_{DSD}$ using Eq.(14)
 }
  \Return{IEC and DSD with parameters $\Theta_c$ and $\Theta_d$}
\label{alg:pseudo_code_training} 
\end{flushleft} 
\end{algorithm}

\noindent\textbf{Dynamic Autoencoder.}
After obtaining the parameter shift of the static autoencoder, the parameters for the encoder and decoder of the dynamic autoencoder are as follows:
\begin{equation}
\begin{aligned}
\Theta_d=
 \begin{cases} 
\Theta_d^E =\Theta_s^E+\Delta\Theta_d^E \\
\Theta_d^D=\Theta_s^D+\Delta\Theta_d^D
\end{cases} 
\end{aligned}
\end{equation}

where $\Theta_d^E$ and $\Theta_d^D$ are the parameters of the dynamic autoencoder, which can adapt the weight of SCD by measuring the parameter shift, thereby improving the OAD accuracy by dynamically modeling the new concepts in the evolving data stream.

Based on the learned parameters of the dynamic autoencoder, given $\vec{x}$, the reconstructing procedure is similar to the static autoencoder:
 \begin{align}
    \vec{x} \Rightarrow \mathcal{E}_d(\vec{x};\Theta_d^E)=\vec{z'} \Rightarrow  \mathcal{D}_d(\vec{z'};\Theta_d^D)=\vec{y'}, \quad s.t. \quad \vec{x} \thickapprox \vec{y'}
\end{align}
where $\vec{z'}$ and $\vec{y'}$ are the latent vector and the reconstructed representation using the dynamic autoencoder respectively.


\begin{algorithm}[t]
\small
\renewcommand{\algorithmicrequire}{\textbf{Input:}}  
\renewcommand{\algorithmicrequire}{\textbf{Initialization:}}

\begin{flushleft}
  \caption{\blue{\name Inference}}
    \textbf{Input}: 
    Evolving stream data $\mathcal{X}$ \\
    \textbf{Output}: Anomaly scores of $\mathcal{X}$\\
    \textbf{Initialization}: Trained SCD, IEC and DSD parametrized by $\Theta_s$, $\Theta_c$  and $\Theta_d$\\

     \For {$\vec{x}$ in $\mathcal{X}$}
     {
     Computing the concept uncertainty $\mathcal{U}_{ctr}(\vec{x})$ using Eq.(6)

       \blue{ \If {$\mathcal{U}_{ctr}(\vec{x})$>$\mu_e$}{ Update SCD to DSD using Eq.(9)\\
        \Return{Anomaly score $L_2(\vec{x},\vec{y'})$}
        }
        
      \Else
     {
        \Return{Anomaly score $L_2(\vec{x},\vec{y})$}
     }
     }}
    \If {\blue{$\sum_{i=t}^{t+\Delta L}({\mathds{1}}_{(\mathcal{U}_{ctr}(\vec{x}_i)>\mu_{e})}  \cdot \mathcal{U}_{ctr}(\vec{x}_i)>\mu_{o} \,\, \mathrm{or} \ \Delta t>T_{max}$ }}
    {
    $\mathcal{X}_h \Rightarrow \mathcal{X}_s$
    
    Update \name
    
    Break
    }    
    
\label{alg:pseudo_code_inference}
\end{flushleft}
\end{algorithm}

\subsubsection{Offline Updating Strategy}
\label{sec:ous}
The entire framework automatically updates its modules based on the accumulated concept uncertainty within a sliding window over the evolving data stream.
Specifically, \name keeps monitoring if the concept uncertainty accumulated within this window surpasses a predetermined threshold, and frequent occurrence of such events indicates that the SCD trained on the historical data streams is incapable of handling the current data stream due to increased concept drift over time.
\blue{This suggests that the entire framework should be updated, namely the SCD, IEC and DSD module, so as to better adapt to the current data stream.}
The Offline Updating Strategy (OUS) is as follows:


\begin{equation}
\begin{aligned}
\!\!\!\!\mathcal{UP}=
 \begin{cases} 
1, \!\! & \mbox{if } \sum_{i=t}^{t+\Delta L}({\mathds{1}}_{(\mathcal{U}_{ctr}(\vec{x}_i)>\mu_{e})}  \cdot \mathcal{U}_{ctr}(\vec{x}_i))>\mu_{o}\,\, \mathrm{or} \ \Delta t>T_{max} \\
0, \!\! & \mbox{else }
\end{cases} 
\end{aligned}
\end{equation}
\noindent
where $\mathcal{UP}$=1 indicates the framework should be updated, $\Delta L$ is the interval of the sliding window, \blue{$\Delta t$} is the time since the last framework update, and $\mu_o$ and $T_{max}$ are the threshold of the framework update and maximum interval since the last update respectively.
$\sum_{i=t}^{t+\Delta L}({\mathds{1}}_{(\mathcal{U}_{ctr}(\vec{x}_i)>\mu_{e})}  \cdot \mathcal{U}_{ctr}(\vec{x}_i)$ aggregates the concept uncertainty greater than the threshold $\mu_{e}$ within the current sliding window.
\blue{We set the threshold $\mu_{e}$ to the largest value of the concept uncertainty during training and update it using an exponential moving average (EMA) strategy~\cite{klinker2011exponential} during updating.
Specifically, we fine-tune the model using data within the current sliding window.
Also, we adopt a parallel training strategy that performs online OAD inference using the latest fine-tuned modules and meanwhile, fine-tunes key modules offline, which decouples the training and inference for efficiency and modularity, and allows for executing OUS without affecting the online OAD inference service.
}

\subsection{Optimization}
\label{sec:optimization}

\name is trained in two stages, and Algorithm~\ref{alg:pseudo_code_training} presents the learning pseudocode: (i) in the first stage, we train the Static Concept-aware Detector (SCD) using the historical data stream $\mathcal{X}_h$.
We note that the initial training of SCD only uses a small subset of the data stream.
The reconstruction error can be computed by taking the $L_2(\cdot)$ loss, \emph{i.e.},
the squared difference between the input and the reconstructed output.
Given $\vec{x}$ in $\mathcal{X}_h$, the LCD loss $\mathcal{L}_{SCD}$ is: 

 \begin{align}
    \mathcal{L}_{SCD}(\Theta_s)=L_2(\vec{x}, \vec{y})=\frac{\sum_{i=1}^{n} (\vec{x}_i-\vec{y}_i)^2}{n}
\end{align}
where $n$ is the dimension of features of the input instance.
(ii) In the second stage, the historical instances are first labeled following Eq.(4), then we follow~\cite{sensoy2018evidential} to train the Intelligent Evolution Controller (IEC).
Specifically, we treat $Dir(\vec{p}|\alpha)$ as a prior on the likelihood and obtain the negated logarithm of the marginal likelihood $\mathcal{L}_{IEC}$ by integrating out the class probabilities:
 \begin{align}
    \mathcal{L}_{IEC}(\Theta_c)=\sum_{c=1}^C({\rm log}(\sum_{c=1}^C \alpha_c)-{\rm log} \alpha_c)\,,
\end{align}

\noindent
The training process of the Dynamic Shift-aware Detector (DSD) is similar to the SCD. Notably, the gradients backpropagated to the hypernetwork together with the SCD.
$\mathcal{L}_{DSD}$ is defined as below: 
 \begin{align}
    \mathcal{L}_{DSD}(\Theta_d)=L_2(\vec{x}, \vec{y'})=\frac{\sum_{i=1}^{n} (\vec{x}_i-\vec{y'}_i)^2}{n}
\end{align}

\noindent
After the two-stage training, \name can perform inference for the incoming data stream, which is summarized in Algorithm~\ref{alg:pseudo_code_inference}.

\subsection{Analysis and Discussion}
\label{sec:analysis_and_discussion}

\noindent
\highlight{Effectiveness.} \revise{\name addresses the concept drift challenge by integrating two detectors and an IEC.
The SCD detector leverages historical data and prior knowledge for the detection, while the DSD detector dynamically learns the parameter shift to enhance SCD and adapts to new concepts effectively in an instance-aware manner.
Notably, IEC determines whether concept drift occurs, circumventing the risk of employing an ineffective detection model for anomaly detection.
With IEC, the adaptability and generalizability of \name are substantially improved, leading to enhanced accuracy in anomaly detection.
In addition, the offline update strategy provides an efficient way to keep up with new concepts in evolving data streams, thereby ensuring high-quality detection results.} 

\noindent
\highlight{Efficiency.} \revise{Efficiency is an important consideration in OAD due to the need for timely responses to evolving data streams.
\name introduces several strategies to achieve high detection efficiency.
\name employs the uncertainty estimate derived from evidential deep learning to detect concept drift, and thus avoids the need for frequent model retraining.
This considerably reduces computational costs and enhances the responsiveness and overall efficiency of the detection model.
\blue{In addition, \name dynamically switches between the base detection model 
and a dynamic shift-aware detector (DSD) supported by the hypernetwork for detecting anomalies.
DSD dynamically generates parameter shifts to account for the current concept, thereby handling concept drift on the fly and enhancing the predictive performance of the base detection model without further fine-tuning or training.} 
The dynamic concept adaptation technique enables METER to harness the strengths of both models, which supports efficient anomaly detection in rapidly changing data streams.
}

\noindent
\highlight{Interpretability.} 
\revise{\name supports interpretability for OAD in terms of providing reliable uncertainty estimates for detection results, which is important and necessary for users in high-stakes applications~\cite{chen2022adaptive,zhao2021empirical,zhao2021identifying}.
We note that softmax probabilities produced by detection models of existing OAD approaches are unreliable uncertainty estimates~\cite{ng2011dirichlet,sensoy2018evidential}.
As such, providing a reliable measure of prediction results is challenging, and is often neglected by existing OAD approaches~\cite{bhatia2022memstream,yoon2021multiple,yoon2022adaptive,kieu2019outlier}.
\blue{
Inspired by recent research utilizing subjective logic (SL) theory~\cite{jsang2018subjective} to improve the interpretability of decision-making processes, e.g., in domains such as multi-view classification~\cite{han2022trusted,gao2023reliable} and molecular property prediction~\cite{soleimany2021evidential},
we derive high-quality uncertainty modeling via evidential deep learning, which utilizes SL theory to explicitly model the reliability of predictions generated by \name.
Specifically, SL formalizes Dempster-Shafer Evidence Theory’s notion of belief assignments over a frame of discernment as a Dirichlet distribution~\cite{sensoy2018evidential}, forming opinions for anomaly detection.
In practice, 
}\name monitors whether concept drift occurs for the current input instance using concept uncertainty modeled by IEC, which can be visualized on a per-input basis for improving the user's understanding of the detection results.
}


\section{Experiments}
\label{sec:experiment}

In this section, we conduct experiments to systematically evaluate the effectiveness, efficiency and interpretability of \name.

\vspace{-1mm}
\subsection{Experimental Setup}
\label{sec:exp_setup}

\subsubsection{Datasets and Applications}\label{sec:dataset}

We adopt \blue{17} real-world benchmark datasets from various domains with different types of concept drift, dimensions, number of data points, and anomaly rates.
These datasets are widely benchmarked in related studies and are representative of evaluating the effectiveness of different OAD approaches to detect anomalies and adapt to concept drift.
Also, four synthetic datasets are introduced to simulate different types and duration of concept drift following the settings in~\cite{yoon2022adaptive}. 
Given the diverse types of data encountered in data streams for OAD, we categorize the datasets into two settings: discrete and continuous.
\blue{The continuous setting is characterized by data streams with temporal dependencies between time steps, while the discrete setting has no or unknown dependencies between time steps.}
By conducting evaluations on both settings, we aim to comprehensively evaluate the performance of different OAD approaches.
The data statistics are summarised in Table~\ref{tab:dataset_stats}. 


\noindent
\textbf{Real-world datasets:}
(1) We first adopt four commonly used anomaly detection datasets from the UCI repository 
and ODDS library~\cite{Rayana:2016}, namely Ionosphere (Ion.), Pima, Satellite, Mammography (Mamm.). The Ionosphere contains anomalies in the radar echoes of the ionosphere. 
The Pima contains information about Pima Indian women who have been tested for diabetes. 
The Satellite contains multi-spectral values of pixels in 3x3 neighborhoods in a satellite image, which is used to identify land as either "barren" or "not barren".
The Mammography dataset consists of "benign" and "malignant" breast X-ray images.
\revise{(2) Secondly, we utilize the BGL~\cite{4273008} 
dataset, a large public dataset consisting of log messages collected from a BlueGene/L supercomputer system at Lawrence Livermore National Labs.
To facilitate analysis, each log message is processed into the structured data format.}
(3) The third category is popular multi-aspect datasets of intrusion detection, namely
KDDCUP99~\cite{kdd99-web} (KDD99)
and NSL-KDD~\cite{tavallaee2009detailed} (NSL). The KDD99 contains network connection records with normal and attack behavior. The NSL is an improvement on the KDD99 that addresses issues such as duplicate data and imbalanced class distribution.
(4) \blue{The next category is time-series datasets procured from two benchmarks: Numenta anomaly detection benchmark~\cite{ahmad2017unsupervised} (NAB) and HexagonML~\cite{UCRArchive2021} (UCR).
We adopt datasets including NYC taxicab (NYC), CPU utilization (CPU), Machine temperature (M.T.) and Ambient temperature (A.T.) from NAB, commonly employed for evaluating streaming anomaly detection algorithms. 
As for UCR, we selectively adopt datasets obtained from natural sources, specifically EPG and ECG.}
(5) We also adopt real-world streaming datasets INSECTS~\cite{souza2020challenges} for simulating concept drift, consisting of optical sensor values collected during monitoring flying insects, with temperature level as the controlled concept. 

\noindent
\textbf{Synthetic datasets:}
Four synthetic datasets~\cite{lecun1998gradient,xiao2017fashion}
are introduced to simulate complex anomaly detection scenarios and data streams~\cite{yoon2022adaptive}. It randomly sets categories as anomaly targets to simulate concepts and sets the duration of each concept randomly to simulate two types of concept drift: "abrupt and recurrent" and "gradual and recurrent".

\begin{table}[t!]
    \small
    \centering
  \renewcommand{\arraystretch}{0.9}
    \caption{\blue{Dataset statistics.}
    }\vspace{-2mm}
    \label{tab:dataset_stats}
    \resizebox{0.95\columnwidth}{!}{
    \begin{tabular}{c|c|cccrc}
    \toprule[1.8pt]
    \multicolumn{1}{c|}{Scenarios}  &\multicolumn{1}{c|}{Settings}&  Datasets     &   $\#$Obj.   &   $\#$Dim.  &   $\#$Outliers ($\%$)     & Concept drift type  \\ 
    \Xhline{1pt}
    \multicolumn{1}{c|}{\multirow{17}{*}{Real}} & \multicolumn{1}{c|}{\multirow{6}{*}{\makecell[c]{Discrete \\ setting}}} & Ionosphere   &   351    &   33      &   126 (35.90$\%$) & Unknown \\
    & & Pima   &   768    &   8    &  268 (34.90$\%$)  & Unknown \\
   & & Satellite   &   6435   &   36   &   2036 (31.64$\%$)& Unknown \\
   & & Mammography   &   11,183     &   6    &   250 (2.32$\%$)    & Unknown\\ 
   & & \revise{BGL}   &  \revise{4,713,493}   &  \revise{ 9}   &  \revise{ 348460 (7.39$\%$)}& Unknown \\
   & & NSL   &  125,973   &   42   &   58630 (46.54$\%$)& Unknown \\
   & & KDD99   &   494,021   &   41   &   97278 (19.69$\%$) & Unknown  \\ \Xcline{2-7}{0.4pt}  
  &  \multirow{10}{*}{\makecell[c]{Continuous \\ setting}} & M.T.  &   22,695    &   10   &   2268 (10.00$\%$)      & Unknown\\
  &  & A.T.  &   7267   &   10   &   726 (10.00$\%$) & Unknown\\
  &  & NYC taxicab  &   10,320     &   10    &  1035 (10.00$\%$)  & Unknown \\
  &  & CPU utilization  &   18,050    &   10   &   1499 (8.30$\%$)  & Unknown  \\
  &  & \blue{EPG} & \blue{30,000} & \blue{10} & \blue{50 (0.17$\%$)} & \blue{Unknown} \\
  &  & \blue{ECG} & \blue{80,000} & \blue{10 }& \blue{250 (0.31$\%$)}  & \blue{Unknown} \\
 \Xcline{3-7}{0.4pt} 
 &   & INSECTS-Abr   &   44,569     &   33      &  529 (1.19$\%$) & Abrupt   \\ 
  &  & INSECTS-Inc   &   48,086    &   33      &   571 (1.19$\%$)  & Incremental    \\ 
 &   & INSECTS-IncGrd   &   20,367    &   33      &    242 (1.19$\%$) & Incremental/gradual  \\ 
 &   & INSECTS-IncRec   &   67,455     &   33    &   800 (1.19$\%$)   & Incremental/recurrent \\ 
      \midrule[1pt]
   \multirow{4}{*}{Synthetic}& \multirow{4}{*}{\makecell[c]{Discrete \\ setting}} & SynM-AbrRec     &   20,480     &   784      &    196 (0.96$\%$)   & Abrupt/recurrent   \\ 
  &  & SynM-GrdRec  &   20,480    &  784       &   192 (0.94$\%$)  & Gradual/recurrent \\ 
 &   & SynF-AbrRec      &   20,480   &  784       &    193 (0.94$\%$) & Abrupt/recurrent   \\
  &  & SynF-GrdRec     &   20,480   &   784    &    204 (1.00$\%$) & Gradual/recurrent  \\ 
    \bottomrule[1.8pt]
    \end{tabular}
    }
    \vspace{-4mm}
\end{table} 

\subsubsection{Baseline Methods}\label{sec:baseline}
We compare \name with 15 baselines in three categories:
(1) Representative and widely used anomaly detection algorithms, namely Local Outlier Factor (LOF)~\cite{breunig2000lof}, Isolation Forest (IF)~\cite{liu2008isolation}, and k-Nearest Neighbors (KNN). STORM~\cite{angiulli2007detecting} is also an important stream data anomaly detection work that is often used as a baseline~\cite{bhatia2022memstream,salehi2016fast,tran2016distance};
(2) Incremental learning-based approach, namely RRCF~\cite{guha2016robust}, 
MStream~\cite{bhatia2021mstream}, and MemStream~\cite{bhatia2022memstream};
(3) Ensemble-based approach, namely HS-Trees~\cite{tan2011fast}, iForestASD~\cite{ding2013anomaly}, RS-Hash~\cite{sathe2016subspace}, LODA~\cite{pevny2016loda}, Kitsune~\cite{mirsky2018kitsune}, xStream~\cite{manzoor2018xstream}, PIDForest~\cite{gopalan2019pidforest}, and ARCUS~\cite{yoon2022adaptive}.
We briefly introduce these baselines as follows.

\powerpoint{LOF}~\cite{breunig2000lof} is a density-based algorithm that measures the local deviation of a point with respect to its neighbors. \powerpoint{IF}~\cite{liu2008isolation} isolates an outlier by recursively partitioning data into smaller subsets using decision trees. \powerpoint{KNN} is a distance-based algorithm that identifies outliers as points with a small number of neighbors within a specified distance. \powerpoint{STORM} ~\cite{angiulli2007detecting} is a distance-based method, which employs a sliding window approach to constantly monitor the streaming data and detect noteworthy deviations from the window's average behavior. \powerpoint{RRCF} ~\cite{guha2016robust} constructs a forest of multiple random trees from data to detect anomalies. 
\powerpoint{MStream}~\cite{bhatia2021mstream} utilizes locality-sensitive hash functions (LSH) to identify unusual group anomalies in data streams. \powerpoint{MemStream}~\cite{bhatia2022memstream}  captures the trend of stream data using a denoising AE with a First-In-First-Out (FIFO) memory module.  \powerpoint{HS-Trees}~\cite{tan2011fast} is a one-class-based ensemble model that can quickly build its tree structure based only on the dimensionality of the data space. \powerpoint{iForestASD}~\cite{ding2013anomaly} integrates sliding windows with Isolation Forest to handle anomaly detection for streaming data. \powerpoint{RS-Hash}~\cite{sathe2016subspace} 
employs an ensemble of subspace grids and randomized hashing, and uses frequency-based scores to detect anomalies in data stream. \powerpoint{LODA} ~\cite{pevny2016loda} an ensemble model of weak detectors such as one-dimensional histograms.
\powerpoint{Kitsune} ~\cite{mirsky2018kitsune} employs an ensemble of AEs to collectively distinguish abnormal patterns. \powerpoint{xStream} ~\cite{manzoor2018xstream} is a density-based ensemble model that performs outlier detection by density estimation on a low-dimensional projection of data points. \powerpoint{PIDForest}~\cite{gopalan2019pidforest} is a random forest-based ensemble model, where each tree is trained on a subset of the data using a partial identification (PID) method to select the most informative features. \powerpoint{ARCUS}~\cite{yoon2022adaptive} employs a model pool to dynamically update its members to detect concept drift.

\begin{table*}[ht]
    \small
    \centering
    \caption{Overall performance comparison of unknown drifts in a discrete setting. A larger score has better performance. Acronym notations of baselines can be found in Sec.~\ref{sec:baseline}. We mark best (bold and underline) and second best(bold) in each row.}\vspace{-2mm}
    \label{tab:overall_resutls_un_dis}
    
    \resizebox{2.1\columnwidth}{!}{
        \begin{tabular}{ c c c c c c c c c c c c c c c c c}
    
    \toprule[1.5pt]
    \multirow{2}{*}{Model Class} & \multirow{2}{*}{Model} & \multicolumn{2}{c}{Ion.}  &   \multicolumn{2}{c}{Pima}  &   \multicolumn{2}{c}{Satellite} & \multicolumn{2}{c}{Mamm.} & \multicolumn{2}{c}{\revise{BGL}}  & \multicolumn{2}{c}{NSL}  &   \multicolumn{2}{c}{KDD99} & Average \\ 
    &   &   AUCROC   &  AUCPR  & AUCROC   &  AUCPR &  AUCROC   &  AUCPR  & AUCROC   &  AUCPR  & AUCROC   &  AUCPR & AUCROC   &  AUCPR & AUCROC   &  AUCPR & Rank \\
    
    \midrule[0.5pt]
  \multirow{4}{*}{Traditional} &  LOF~\cite{breunig2000lof}  & 
0.874 & 0.827  & 0.542 & 0.371 & 0.598 & 0.481 & 0.720 & 0.089 & 0.542 & 0.206 & 0.586 & 0.428 & 0.653 & 0.359 & 10.133\\
  &  IF~\cite{liu2008isolation}  &  0.860 & 0.817 &  0.677 &  0.502 &  0.676 & 0.375 &  0.867  & 0.211 &  \textbf{0.823} & 0.295 & 0.530 & 0.577 & 0.784 & 0.406 & 7.333\\
   &  KNN  &  \textbf{0.929} & \textbf{0.932} &0.615 & 0.457 & 0.677 & 0.539 & 0.839 & 0.156 & 0.765 & 0.274 & 0.897 &  0.899 & 0.946 & 0.902 & 6.133 \\
  &  STORM~\cite{angiulli2007detecting} & 0.640 & 0.526 & 0.529 & 0.373 & 0.680 & 0.452 & 0.615 & 0.418 & 0.203 & 0.043 & 0.513 &  0.138 & 0.913 &  0.822 & 10.933 \\

    \midrule[0.5pt]
    \multirow{4}{*}{Incremental} &  RRCF~\cite{guha2016robust}  & 0.586 & 0.411 & 0.575 & 0.393 & 0.553 & 0.356 & 0.713 & 0.524 & 0.540 & 0.076 & 0.604 & 0.534 & 0.773 & 0.347 &  11.067 \\
    &   MStream~\cite{bhatia2021mstream}  & 0.681 & 0.486 & 0.524 & 0.440 & 0.647 & 0.457 & 0.798 &	0.076 & 0.531 & 0.105 & 0.759 & 0.716 & 0.958 & \textbf{0.912} & 9.333 \\
         
    &   MemStream~\cite{bhatia2022memstream}    &   0.821  &  0.672 &   \textbf{0.703}  & 0.551   &  0.722 & \textbf{0.682} & \textbf{0.902} & 0.225 & 0.694 & 0.144 & \textbf{\underline{0.988}}  &   \textbf{\underline{0.967}} & \textbf{0.979} &   0.857 & 5.200 \\

    \midrule[0.5pt]
    \multirow{8}{*}{Ensemble} &   HS-Trees~\cite{tan2011fast}  & 0.687 & 0.574 & 0.667 & 0.344 & 0.512 & 0.348 & 0.797 & 0.623 & 0.599 & 0.174 & 0.806 & 0.735 & 0.901 & 0.728 & 9.267 \\
    &   iForestASD~\cite{ding2013anomaly}  &  0.744 & 0.601 & 0.515 & 0.356 & 0.642 & 0.451 & 0.575 & 0.031 & 0.701 & \textbf{\underline{0.382}} &  0.511 & 0.483 & 0.532 & 0.227 & 11.000 \\
   &   RS-Hash~\cite{sathe2016subspace}    & 0.743 & 0.502 & 0.518 & 0.372 & 0.640 & 0.586 & 0.776 & 0.622 & 0.436 & 0.245 & 0.684   & 0.524 &  0.783 & 0.707 & 10.000 \\
    &   LODA~\cite{pevny2016loda}   & 0.514 & 0.373 & 0.501 & 0.347 & 0.500 & 0.316 & 0.500 & 0.023 & 0.523 & 0.074 & 0.504 & 0.535  & 0.507 & 0.197 & 14.000\\    
    &   Kitsune~\cite{mirsky2018kitsune}   & 0.920 & 0.896 & 0.590 & 0.451 & 0.732 & 0.673 & 0.603 & 0.202 &	0.514 & 0.074 & 0.947 & 0.918 & \textbf{\underline{0.982}} & \textbf{\underline{0.993}} & 6.533\\ 
    &   xStream~\cite{manzoor2018xstream}  & 0.773 & 0.591 & 0.656 & \textbf{0.583} & 0.659 & 0.533 & 0.847 & \textbf{\underline{0.630}} & 0.623 & 0.356 & 0.540 & 0.327 & 0.954 & 0.881 & 7.067 \\
    &  PIDForest~\cite{gopalan2019pidforest}   & 0.821 & 0.718 & 0.669 & 0.474 & 0.718 & 0.543 & 0.847 & 0.202 & 0.791 & 0.300 & 0.503 & 0.561 & 0.864 & 0.772 & 7.467\\
    &   ARCUS~\cite{yoon2022adaptive}    & 0.919 & 0.894 & 0.607 & 0.420 & \textbf{\underline{0.797}} & 0.560 & 0.812 & \textbf{0.261} & 0.768 & 0.185 & 0.262 & 0.365 & 0.972 & 0.807 & 7.533\\
    
  \midrule[0.5pt]
  
   \rowcolor{gray!40} Ours   &   \name  &  \textbf{\underline{0.950}}  &  \textbf{\underline{0.956}}  &     \textbf{\underline{0.733}}  &  \textbf{\underline{0.654}}  & \textbf{0.796}  &  \textbf{\underline{0.777}}  &   \textbf{\underline{0.913}}  &  0.491  &  \textbf{\underline{0.895}} & \textbf{0.369} & \textbf{0.982}  &  \textbf{0.963}  & 0.973  &  0.853 & 3.0001 \\
    
    \bottomrule[1.5pt]
        \end{tabular}
    } 
\end{table*}

\subsubsection{Evaluation Metrics} \label{sec:metrics}

We adopt AUCROC and AUCPR as evaluation metrics.
AUCROC is the area under ROC curve, which plots the false-negative rate (FNR) as the x-axis and the true-positive rate (TPR) as the y-axis at different thresholds.
AUCPR is the area under PR curve, which plots the precision against recall at different thresholds.
The metrics fall within the range $[0, 1]$, and a higher value indicates better detection performance.


\subsubsection{Implementation Details} \label{sec:implementation_details}
We implement LOF and IF using the \emph{scikit-learn} library~\cite{pedregosa2011scikit}, and KNN using the pyod library~\cite{zhao2019pyod}.
The open-source PySAD library~\cite{yilmaz2020pysad} is used to implement STORM~\cite{angiulli2007detecting}, RRCF~\cite{guha2016robust}, HS-Trees~\cite{tan2011fast}, iForestASD~\cite{ding2013anomaly}, RS-Hash~\cite{sathe2016subspace}, LODA~\cite{pevny2016loda}, and xStream~\cite{manzoor2018xstream} with default parameters.
For other baseline methods such as MStream~\cite{bhatia2021mstream}, MemStream~\cite{bhatia2022memstream}, Kitsune~\cite{mirsky2018kitsune}, PIDForest~\cite{gopalan2019pidforest}, ARCUS~\cite{yoon2022adaptive}, we adopt the official implementations, using the recommended parameter settings.
For ARCUS, we use the base model RAPP~\cite{kim2020rapp}.
In cases where default parameter values are not provided, we conduct a grid search to find the optimal parameters that yield the best performance.
Adam~\cite{adam} is used as an optimizer in all learning-based anomaly detection models with a learning rate searched in $0.1 \unsim 1e$-3.

For \name, the encoder and decoder are implemented as 2 to 10-layer DNNs with a symmetric structure, where the autoencoder's latent space dimension is set to the number of principal components to ensure a minimum of 70\% explained variance following prior research~\cite{yoon2022adaptive,kim2020rapp}. 
IEC is a two-layer DNN with the ReLU activation function.
We use an Adam optimizer with a learning rate of $1e$-2 with an exponential decay rate of 0.96. The number of epochs is set to 1000.
We utilize grid search for hyperparameter tuning.
Specifically, the threshold rate $\mu_{p}$ for the pseudo labels from SCD is searched within the range of 0.05 to 0.5 with an interval of 0.05.
The threshold of the concept uncertainty $\mu_{e}$ is searched in \{0.001, 0.005, 0.01, 0.1, 0.2, 0.4\}.
While the threshold of the offline updating strategy $\mu_{o}$ is in the range of 0.1 to 1 of the window size, and the window size $\Delta L$ is set to 64.
\blue{The historical data ratio $h_r$ is set to 0.2, meaning that 20\% of the dataset is utilized as a historical data stream $\mathcal{X}_s$ for training purposes.
We conduct a sensitivity analysis on the three key threshold hyperparameters of our framework, along with the window size $\Delta L$ and the historical data ratio $h_r$ in Section~\ref{sec:exp_sensitivity}. We create a data generator to simulate the generation of streaming data and
}
report the average value of 5 independent runs for all baselines.


\blue{All the experiments are conducted in a server with Xeon(R) Silver 4114 CPU @ 2.2GHz (10 cores), 256G memory, and GeForce RTX 2080 Ti.
All the models are implemented in PyTorch 1.10.0 with CUDA 10.2.}

\begin{figure}[t]
    \centering
    \includegraphics[width=0.9\linewidth]{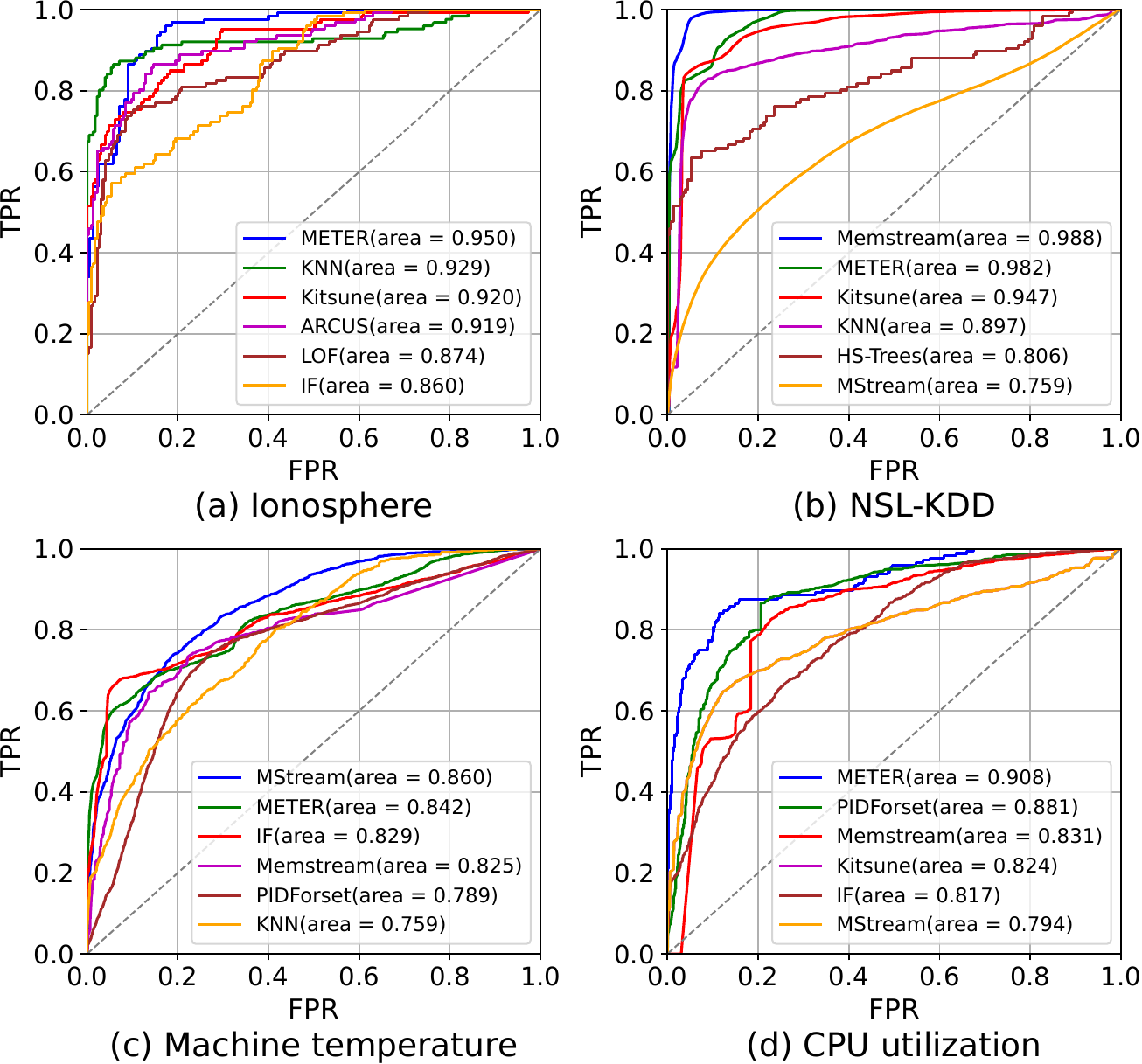} \vspace{-1mm}
    \caption{ \revise{Receiver operating characteristic curves of the top six methods on real-world datasets.
    }}
    \label{fig:auc}
     \vspace{-2mm}
\end{figure}

\subsection{Effectiveness}
\label{sec:exp_effectiveness}

\vspace{1mm}
\noindent
\powerpoint{\revise{Comparison on real-world benchmark datasets.}} We compare \name with other baselines on the \blue{17 real-world datasets} as summarized in Table \ref{tab:dataset_stats}, with results reported in Table \ref{tab:overall_resutls_un_dis}, \ref{tab:overall_resutls_un_con} and \ref{tab:overall_resutls_kn_con}.
Our \name achieves the best performance in most scenarios, across different concept drift types and problem settings (discrete or continuous). 
Significant improvements are observed in terms of AUCROC in Table \ref{tab:overall_resutls_un_dis}, e.g. 2.2\% on Ionosphere and 4.3\% on Pima, which are consistent with the results illustrated in Figure~\ref{fig:auc}.
\blue{Meanwhile, \name obtains the highest AUCROC of 0.968 and the second-highest AUCROC of 0.778 (very close to the best AUCROC of 0.780 achieved by MemStream on ECG) on the more challenging dataset EPG and ECG, respectively.}
\blue{Likewise, drawing insights from Table~\ref{tab:overall_resutls_un_con}, \name performs exceptionally well on time series data, which is the top-ranked model on average.
Notably, for preprocessing and better modeling time series data, we follow ~\cite{gopalan2019pidforest} and adopt the shingling technique with a window width of 10.
Then, these transformed vectors are supplied to \name, equipping it with the ability to capture short-term temporal dependencies within the time series window. Furthermore, the hypernetwork-based DSD adapts to long-term variations within the time series by learning the nuances in SCD weights. This adaptive mechanism endows \name with the capability to discern and respond to concept shifts in data patterns.}

While some methods demonstrate remarkable performance on certain datasets, they lack consistency.
\blue{For instance, ARCUS achieves the highest AUCPR on EPG, while it only obtains an AUCPR of 0.340, i.e., 34.99\% worse than STORM with the highest AUCPR of 0.523.}
\blue{Table~\ref{tab:overall_resutls_kn_con} illustrates that about half of the methods exhibit subpar performance on real-world datasets with known concept drift. 
Notably, although methods like MemStream obtain competitive performance on INSECT-Abr and INSECT-IncRec, they perform worse on the other two datasets, whose performance is even worse than \name using only the static concept-aware detector as shown in Table~\ref{tab:ablation}. This demonstrates the complexity of OAD when dealing with real-world datasets characterized by distinct concept drifts. Different kinds of concept drifts require very different modeling strategies, and thus resilient models that can adapt to a wide spectrum of concept drift scenarios are much needed.}
\blue{In this context, consistently outperforms baseline models across various settings and types of concept drift, achieving overall the highest rank across datasets.} 
Also, \name shows high computational efficiency.
As shown in Table \ref{tab:overall_resutls_kn_con}, the average running time of \name on INSECTS is only 88s, which is substantially lower than most of the baselines, while still delivering superior performance.

\begin{table*}[ht]
    \small
    \centering\vspace{-2mm}
    \renewcommand{\arraystretch}{0.7} \vspace{-2mm}
    \caption{\blue{Overall performance comparison of Unknown drifts in a continuous setting.}}\vspace{-2mm}
    \label{tab:overall_resutls_un_con}
    
    \resizebox{2.1\columnwidth}{!}{
        \begin{tabular}{ c c c c c c c c c c c c c c c}
    
    \toprule[1.5pt]
    \multirow{2}{*}{Model Class} & \multirow{2}{*}{Model} & \multicolumn{2}{c}{M.T.}  &   \multicolumn{2}{c}{A.T.}  &   \multicolumn{2}{c}{NYC.}  &   \multicolumn{2}{c}{CPU.} &\multicolumn{2}{c}{\blue{EPG}} &\multicolumn{2}{c}{\blue{ECG}} &  \multirow{2}{*}{\blue{Average Rank}}  \\
    &   &   AUCROC   &  AUCPR  &  AUCROC   &  AUCPR &  AUCROC   &  AUCPR  & AUCROC   &  AUCPR  & \blue{AUCROC} & \blue{AUCPR}& \blue{AUCROC} &  \blue{AUCPR}  & \\
    
    \midrule[0.5pt]
  \multirow{4}{*}{Traditional} &  LOF~\cite{breunig2000lof}  & 0.501 & 0.141 &  0.563 & 0.126 &  0.671 &  0.211 & 0.560 &  0.112 & \blue{\textbf{0.934}}&\blue{0.679} & \blue{0.670}&\blue{0.016}  &\blue{9.667}  \\
  &  IF~\cite{liu2008isolation}  & 0.829 & 0.573 & 0.762 & 0.362 & 0.624 & \textbf{0.331} & 0.817  & \textbf{\underline{0.760}} & \blue{0.811} & \blue{0.552}&  \blue{0.668} & \blue{0.005} & \blue{5.500} \\
   &  KNN  & 0.759 & 0.255 &  0.634 & 0.200  &  \textbf{0.697} & 0.202 & 0.724 & 0.452 & \blue{0.083} & \blue{0.001}&  \blue{0.247} &\blue{0.002}& \blue{10.000}\\
  &  STORM~\cite{angiulli2007detecting}  & 0.604 & 0.127 & 0.518 &  0.105 & 0.460 & 0.097 & 0.667 & 0.605 & \blue{0.578} &\blue{0.436} & \blue{0.662}&\blue{\textbf{\underline{0.523}}}& \blue{11.667} \\
    \midrule[0.5pt]
    \multirow{4}{*}{Incremental} &   RRCF~\cite{guha2016robust}  &  0.628 & 0.153 & 0.519 & 0.110 & 0.502 & 0.121 & 0.617 &  0.368 & \blue{0.814} & \blue{0.498} & \blue{0.387} &\blue{0.002} & \blue{11.667}\\
    &   MStream~\cite{bhatia2021mstream}   & \textbf{\underline{0.860}} &	0.505 & 0.619 &	0.156 & 0.639 & 0.168 & 0.794 & 0.443 &	\blue{0.824} &\blue{0.621} & \blue{0.721}&\blue{0.284} &\blue{6.417}\\
    &   MemStream~\cite{bhatia2022memstream}  &  0.825  &   \textbf{0.573 }  & 0.722  &   0.334  & \textbf{\underline{0.731}}  &  0.311 &   0.831  &  0.227  & \blue{0.930} & \blue{0.656} & \blue{\textbf{\underline{0.780}}} & \blue{0.007} &\blue{4.500}\\
    
    \midrule[0.5pt]
    \multirow{8}{*}{Ensemble} &   HS-Trees~\cite{tan2011fast}  & 0.617 & 0.359 & 0.522 & 0.310 & 0.558 &  0.269 & 0.678 & 0.585 &\blue{0.531}&\blue{0.334}& \blue{0.621}&\blue{0.439}& \blue{9.083}  \\
    &   iForestASD~\cite{ding2013anomaly}   & 0.738 & 0.231 & 0.514 & 0.167 & 0.501 & 0.117 & 0.755 & 0.153 & \blue{0.782} &\blue{0.470} & \blue{0.733}& \blue{0.006}& \blue{10.167}\\
    &   RS-Hash~\cite{sathe2016subspace}   & 0.607 & 0.549 & 0.742 & 0.180 & 0.524 & 0.106 & 0.712 &  0.467 & \blue{0.552}& \blue{0.186}& \blue{0.584} &\blue{0.203} &\blue{9.833} \\
    &   LODA~\cite{pevny2016loda}  & 0.503  & 0.100 & 0.499 & 0.101 &  0.499 & 0.101 & 0.500 &  0.083 & \blue{0.595} & \blue{0.182} &\blue{0.721} &\blue{0.077} &\blue{13.500} \\
    &   Kitsune~\cite{mirsky2018kitsune}   & 0.684 & 0.416 & 0.599 & 0.274 & 0.465 & 0.124& 0.824 & 0.669 & \blue{0.897} &\blue{0.020}&  \blue{0.726}&\blue{0.231}&\blue{7.833}  \\ 
   &   xStream~\cite{manzoor2018xstream}   &  0.696 &  0.596 & 0.567 & 0.319 &  0.586 & 0.121  & 0.730 &  0.195 & \blue{0.687} &\blue{0.158} & \blue{0.705}& \blue{0.365} & \blue{8.250}  \\
    &  PIDForest~\cite{gopalan2019pidforest}  &  0.789 & 0.389 & \textbf{\underline{0.797}} & 0.320 & 0.513 & 0.113 &  \textbf{0.881} & 0.439 & \blue{0.836}&\blue{0.238}& \blue{0.683}&\blue{0.006}&  \blue{7.667}\\
    &   ARCUS~\cite{yoon2022adaptive}   &  0.376 & 0.511 & 0.518 & \textbf{0.389} & 0.470 & 0.317 & 0.678 & 0.128 & \blue{0.885} &\blue{\textbf{\underline{0.724}}} & \blue{0.682}&\blue{0.340} & \blue{8.417} \\
  \midrule[0.5pt] 
  \rowcolor{gray!40} Ours    &   \name  & \textbf{0.842} &  \textbf{\underline{0.652}}  &  \textbf{0.782}  &  \textbf{\underline{0.399}}  & 0.688  & \textbf{\underline{0.386}} & \textbf{\underline{0.908}} & \textbf{0.715}& \blue{\textbf{\underline{0.968}}}& \blue{0.625}& \blue{\textbf{0.778}}& \blue{\textbf{0.495}}&\blue{1.833}\\
    
    \bottomrule[1.5pt]
        \end{tabular}
    }
\end{table*}

\begin{table}[t]
    \small
    \centering
    \renewcommand{\arraystretch}{0.8}
    \caption{Overall performance comparison of known drifts in a continuous setting.}\vspace{-2mm}
    \label{tab:overall_resutls_kn_con}
    \resizebox{1\columnwidth}{!}{
        \begin{tabular}{ c c c c c c c c}
    
    \toprule[1.5pt]
    \multirow{2}{*}{Model Class} & \multirow{2}{*}{Model} & INSECTS  &   INSECTS  & INSECTS &  INSECTS & \multirow{2}{*}{Average Rank} & \multirow{2}{*}{Time (s)} \\
    & & -Abr& -Inc& -IncGrd  &-IncRec &  &  \\
    \midrule[0.5pt]
  \multirow{4}{*}{Traditional} &  LOF~\cite{breunig2000lof}  & 0.578 & 0.556 &  0.589 & 0.526 & 11.250 & 180 \\
    &  IF~\cite{liu2008isolation}  &  0.679 & 0.632 & 0.697 & 0.593 & 6.500 & 67 \\
    &  KNN  &  0.666 & 0.597 &  0.673 & 0.553 & 8.000 & 105 \\
&  STORM~\cite{angiulli2007detecting} & 0.408 & 0.441 & 0.446 & 0.449 & 16.500 & 122\\

    \midrule[0.5pt]
    \multirow{4}{*}{Incremental} &   RRCF~\cite{guha2016robust} & 0.600 & 0.579 &0.624 & 0.593 & 9.250 & 121 \\
    &   MStream~\cite{bhatia2021mstream}  & 0.703 & \textbf{0.698} &	\textbf{\underline{0.788}} & \textbf{0.672} & 3.250 & 18\\
    &   MemStream~\cite{bhatia2022memstream}  &  0.753  &  0.348 &   0.728 &  0.361 & 10.750 & 109 \\
    
    \midrule[0.5pt]
    \multirow{8}{*}{Ensemble} &   HS-Trees~\cite{tan2011fast}  & 0.499 & 0.507 & 0.497 & 0.499 & 14.250 & 302\\
    &   iForestASD~\cite{ding2013anomaly} & 0.599 & 0.589 & 0.616 & 0.575 & 9.500 & 7985\\
    &   RS-Hash~\cite{sathe2016subspace}   & 0.484 & 0.509  & 0.459 & 0.506 & 14.250 &225\\
    &   LODA~\cite{pevny2016loda}  &  0.498 & 0.503 &0.496 & 0.499 & 14.750 & 831\\
    &   Kitsune~\cite{mirsky2018kitsune}   &\textbf{ 0.759} & 0.584 & 0.730 & 0.594 & 5.250 & 164 \\ 
    &   xStream~\cite{manzoor2018xstream}  &  0.514 & 0.516 & 0.533 & 0.504 & 12.750 & 408\\
    &  PIDForest~\cite{gopalan2019pidforest}  & 0.757 & 0.675 & \textbf{0.748}& 0.631 & 3.500 & 18047 \\
    &   ARCUS~\cite{yoon2022adaptive}  &  0.601 &   0.597  & 0.576 &   0.632  & 8.000 & 79\\
  \midrule[0.5pt]
\rowcolor{gray!40} Ours    &   \name  &   \textbf{\underline{0.816}} &  \textbf{\underline{0.795}}  & 0.712 &  \textbf{\underline{0.794}}  & 2.250 & 88  \\
    \bottomrule[1.5pt]
        \end{tabular}
    }
     \vspace{-2mm}
\end{table}

\noindent
\powerpoint{\revise{Comparison on Synthetic datasets.}} To further validate the effectiveness of \name on high dimensional data, we conduct experiments on four synthetic datasets, with results summarized in Table \ref{tab:overall_resutls_kn_dis}.
Despite the increased data dimensions, \name still outperforms other baselines.
Although ARCUS occasionally matches or even slightly outperforms \name in AUCROC, its performance is not consistent.
Specifically, ARCUS suffers a significant performance decline on INSECTS compared to \name, i.e., a gap of 35.8\% and 33.2\% on INSECTS-Abr and INSECTS-Inc, respectively.


In these experiments, \name consistently demonstrates superior performance across varied experimental settings, data dimensions, and drift types, showing its excellent adaptability and robustness.
Moreover, \name maintains high computational efficiency while obtaining such remarkable detection performance.

\subsection{Concept Drift}
\label{sec:exp_concept_drift}
To validate the effectiveness of our approach in real-time detection and rapid response to concept drift, we monitor the evolution of concept uncertainty of \name (depicted as a blue line) on real data streams subject to concept drift, specifically on INSECTS.
In this experiment, the changes in temperature are used as indicators of concept drift.
In addition, we plot the timing of model updates (indicated by the orange dotted line) and the corresponding changes in AUCROC (indicated by the gray line) over time.
This tracks the performance of \name as it adapts to evolving data streams.
For a comprehensive understanding of the monitoring and adaptation effects of concept drift, we adopt equidistant sampling across the entire dataset, given that our method operates at the instance level.
By selecting 100 equidistant points for both uncertainty and AUCROC analysis, we ensure a representative snapshot of the data stream's evolution.

\begin{figure*}[t]
    \centering  
     \vspace{-3mm}
    \includegraphics[width=0.95\linewidth]{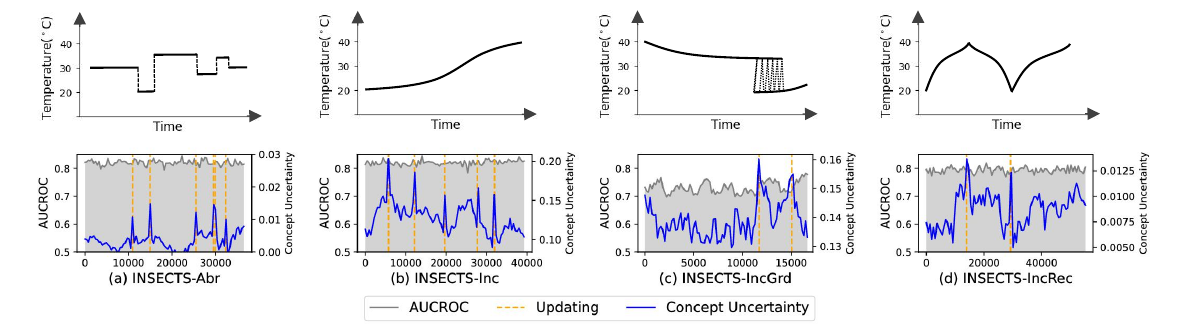} \vspace{-1mm}
    \caption{ Analysis of concept drift adaptation on INSECTS dataset.
    }
    \label{fig:concept_drift}
\end{figure*}

As shown in Figure~\ref{fig:concept_drift}, despite some volatility in uncertainty across the data stream, there is generally a sharp increase in uncertainty when concept drift occurs, with virtually no delay.
\blue{We note that the initial drift results in a sharp spike in the concept uncertainty at the beginning of concept drift, e.g., the time step around 7000 in INSECTS-Inc, and more moderated concept uncertainty during the subsequent concept drifts, even if an increasing trend is observed, such as the concept shift occurring at time step 20000.
This phenomenon is also consistent with the observation in INSECTS-IncGrd and INSECTS-IncRec.
}
Further, the timing of model updates aligns with these sharp increases in uncertainty, corroborating the rapid response of \name to the emergence of anomalies.
Importantly, \name consistently exhibits high AUCROC scores on real data streams, with no significant fluctuations in the presence of concept drift.
This consistency demonstrates the stability and reliability of our approach.

\subsection{Ablation Study}
\label{sec:exp_ablation}

\begin{figure}[t]
    \centering
    \includegraphics[width=0.9\linewidth]{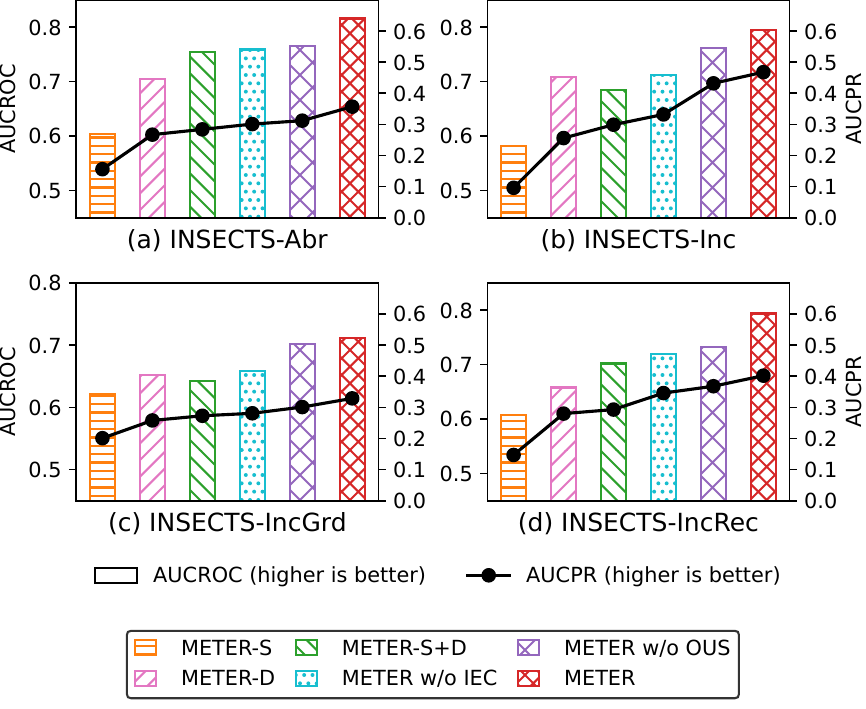}\vspace{-2mm}
    \caption{ Ablation analysis. The AURROC and AUCPR performance of six \name variants on INSECTS datasets.
    }
    \label{fig:ablation_auc}
    \vspace{-2mm}
\end{figure}

\noindent
\blue{\textbf{Effectiveness of each module:} Extensive ablation studies are conducted on four real-world datasets with different types of concept drifts to evaluate the contribution of individual modules in \name. }
The composition of the variants and their average AUCROC results on INSECTS are detailed in Table~\ref{tab:ablation}, with separate results provided in Figure~\ref{fig:ablation_auc}.
The results show that each module contributes considerably to improving the detection performance.
Notably, the experiments suggest that DSD plays a more important role than SCD.
Also, both IEC and OUS, introduced to detect and address concept drift respectively, are critical to support effective OAD.
Similar results can be observed in Figure~\ref{fig:ablation_auc}.
Further, the increase in the detection performance of \name becomes more prominent as the data stream's sample size grows (e.g., the gain curve on INSECTS-IncRec with 67,455 samples is steeper than that on INSECTS-IncGrd with 20,367 samples), demonstrating the scalability of \name to large-scale data stream scenarios.

\begin{table}[t]
    \small
    \centering
    \renewcommand{\arraystretch}{0.8}
    \caption{\revise{Overall performance comparison of known drifts in a discrete setting.}} \vspace{-2mm}
    \label{tab:overall_resutls_kn_dis}
    
    \resizebox{1\columnwidth}{!}{
        \begin{tabular}{ c c c c c c c c }
    
    \toprule[1.5pt]
    \multirow{2}{*}{Model Class} & \multirow{2}{*}{Model} & SynM  &  SynM  &   SynF &  SynF  & \multirow{2}{*}{Average Rank} & \multirow{2}{*}{Time (s)} \\
  &  & -AbrRec & -GrdRec & -AbrRec & -GrdRec &  & \\
  \midrule[0.5pt]
   \multirow{4}{*}{Traditional} &  LOF~\cite{breunig2000lof}  & 0.554 & 0.539 & 0.473 & 0.503 & 11.500 & 620\\
   &  IF~\cite{liu2008isolation}  &  0.565 & 0.544 & 0.497 & 0.478 & 11.250 & 328\\
   &  KNN  &  0.550 & 0.584 & 0.515 & 0.513 & 8.000 & 55 \\
  &  STORM~\cite{angiulli2007detecting} &  0.513 & 0.516 & 0.521 & 0.530 & 9.500 & 323\\   
    \midrule[0.5pt]
    \multirow{4}{*}{Incremental} &   RRCF~\cite{guha2016robust} & 0.695 & 0.666 & 0.681 & \textbf{0.715} & 2.750 & 142\\
    &  MStream~\cite{bhatia2021mstream}  & 0.473 & 0.608 & 0.623 & 0.507 & 9.500 & 31\\
    &  MemStream~\cite{bhatia2022memstream}  &  0.517  &  0.496 & 0.523  &   0.472 & 13.500 & 549 \\
    \midrule[0.5pt]
    \multirow{8}{*}{Ensemble} &   HS-Trees~\cite{tan2011fast} & 0.497 & 0.503 & 0.500 & 0.511 & 12.750 & 139 \\
    &   iForestASD~\cite{ding2013anomaly} &  0.523 & 0.514 & 0.533 &  0.506 & 10.000 & 2580 \\
   &   RS-Hash~\cite{sathe2016subspace}   & 0.482 & 0.506 & 0.470 & 0.506 & 14.500 & 102 \\
    &   LODA~\cite{pevny2016loda}  & 0.504 & 0.500 & 0.502 & 0.500 & 14.500 & 328\\
    &   Kitsune~\cite{mirsky2018kitsune}   & 0.544  & 0.502 & 0.570  & 0.523 & 9.250 & 323 \\ 
    &   xStream~\cite{manzoor2018xstream}  & 0.641 & 0.658  & 0.625 & 0.605 & 4.500 & 997\\
    &  PIDForest~\cite{gopalan2019pidforest}  &  0.514 & 0.546 &  0.513 &  0.503 &11.000 &  8176 \\
    &   ARCUS~\cite{yoon2022adaptive}   & \textbf{\underline{0.901}} & \textbf{\underline{0.894}} & \textbf{0.759} & \textbf{\underline{0.787}} & 1.25 & 120 \\
  \midrule[0.5pt]
\rowcolor{gray!40} Ours    &   \name  &   \textbf{0.879} &\textbf{0.825} & \textbf{\underline{0.806}} & 0.632 & 2.000 & 72 \\
    \bottomrule[1.5pt]
        \end{tabular}
    } \vspace{-2mm}
\end{table}

\begin{table}[t!]
    \small
    \centering
    \renewcommand{\arraystretch}{0.8}
    \caption{ Performance of ablation experiments, where AUC is the average AUCROC value on the four INSECTS datasets. }\vspace{-2mm}
    \label{tab:ablation}
    \resizebox{0.9\columnwidth}{!}{
        \begin{tabular}{ l c c c c c }
    \toprule[1.5pt]
   	Variant & SCD & DSD & IEC & OUS &  AUC \\
		\hline	
		\name-S & \checkmark & $\times$ & $\times$ & $\times$ & 0.604 \\
		\name-D & $\times$ & \checkmark & $\times$  & $\times$ & 0.681 \\
		\name-S+D & \checkmark & \checkmark & $\times$  & $\times$  & 0.696 \\
        \name w/o IEC & \checkmark & \checkmark & $\times$  & \checkmark  & 0.713\\
		\name w/o OUS &  \checkmark & \checkmark &  \checkmark & $\times$ &  0.741 \\	
		\name & \checkmark & \checkmark & \checkmark & \checkmark  &  0.779 \\   
    \bottomrule[1.5pt]
    \end{tabular} 
    }\vspace{-2mm}
\end{table}
\begin{table}[t!]
    \small
    \centering
    \renewcommand{\arraystretch}{0.9}
    \caption{ \revise{\blue{Effects of instance-specific information, prior knowledge and various SCD implementations on \name.}}}\vspace{-2mm}
    \label{tab:ablation_instance-specific}
    \resizebox{1.0\columnwidth}{!}{
        \begin{tabular}{ l c c c c c }
    \toprule[1.5pt]
   	\multirow{2}{*}{Variant} & INSECTS  &   INSECTS  & INSECTS &  INSECTS &  \multirow{2}{*}{Average} \\    
    & -Abr& -Inc& -IncGrd  &-IncRec &   \\
		\hline	
    \name-re & 0.640 & 0.482 & 0.526 & 0.617  &  0.566 \\
    \revise{\name-pl} & 0.827 &  0.806 &  0.762 &  0.782  & 0.794   \\
		\name & 0.814 & 0.795 & 0.713 & 0.795  &  0.779 \\
            \blue{\name-lstm} & \blue{0.915} & \blue{0.844} & \blue{0.765} & \blue{0.827} & \blue{0.838}\\
  		\blue{\name-conv} & \blue{0.893} & \blue{0.825} & \blue{0.693} & \blue{0.816} & \blue{0.807}  \\
      	\blue{Res-\name-conv} & \blue{0.910} & \blue{0.832} & \blue{0.742} & \blue{0.821} & \blue{0.826} \\
    \bottomrule[1.5pt]
    \end{tabular}
    }
\end{table}
\noindent
\blue{\textbf{Instance-specific information:}
A dedicated ablation study is conducted to assess the superiority of adopting the instance-aware input for hypernetwork over the conventional approach of using random embeddings. We denote the variant by \name-re.
Remarkably, instance-specific information significantly improves the detection performance of \name.
In contrast to the original hypernetwork, which uses random embeddings and thus has a weak correlation between parameter generation and the current input instance, \name takes into account instance-specific inputs, leading to instance-aware modeling.
This enables \name to effectively capture the dynamic changes in streaming data, thereby making it more efficient and effective for online anomaly detection.}

\noindent
\blue{\textbf{Prior knowledge:}
We introduce a variant of \name to evaluate its adaptability and generalization in situations with limited labeled samples.
Table~\ref{tab:ablation_instance-specific} reports the experimental results, where \name-pl denotes the pseudo-labeling strategy enhanced by incorporating 1\% of labeled anomalies into the training set.
The results in Table~\ref{tab:ablation_instance-specific} also reveal that the evidential IEC successfully leverages prior knowledge and considerably enhances the learning capacity of \name by incorporating only a small number of labeled samples.}

\noindent
\blue{\textbf{Flexibility:} We conducted comprehensive testing of various SCD designs to thoroughly explore the flexibility of \name, including LSTM, 1D convolution, and 1D convolution with residual connections. For a fair comparison, all the variants share the same base structure as \name's DNNs implementation. 
Results in Table~\ref{tab:ablation_instance-specific} show that all the three enhanced SCD modules can achieve noticeably better detection performance, with an increase in AUCROC of 7.57\%, 3.59\%, and 6.03\% respectively, as compared to the original SCD module based on a canonical DNNs.
This not only confirms the adaptability and flexibility of \name but also points out enhancement directions for further improving \name's performance.
}
\begin{table}[t!]
    \small
    \centering
    \renewcommand{\arraystretch}{0.9}
    \caption{ \blue{Validation of recurring patterns.}}\vspace{-2mm}
    \label{tab:reccurring}
    \resizebox{1.0\columnwidth}{!}{
        \begin{tabular}{ c c c c c c c }
    \toprule[1.5pt]
   	Variant & $p_{1-test}$ & $p_{2-test}$ & $p_{3-test}$ & $p_{4-test}$ & $p_{5-test}$ & Average \\ 
    \hline	
        Group I & 0.804 & 0.681 & 0.711 & 0.702 & 0.801 & 0.740 $\pm$ 0.052 \\
	Group II & 0.804 & 0.710 & 0.734 & 0.754 & 0.837 & 0.768 $\pm$ 0.046\\
    \bottomrule[1.5pt]
    \end{tabular}
    }
    \vspace{-1mm}
\end{table}
\begin{table}[t!]
   \small
    \centering
    \renewcommand{\arraystretch}{0.8}
    \caption{\blue{Training and inference efficiency of \name.}}\vspace{-2mm}
    \label{tab:efficiency}
    
    \resizebox{1.0\columnwidth}{!}{
        \begin{tabular}{c c c c c c}
    \toprule[1.5pt]
    \multirow{2}{*}{Dataset} & \multicolumn{2}{c}{Throughput} & Training & \multirow{2}{*}{Memory(MiB)} \\
   &  Training  &  Inference & Time(s) &    \\
    
    \midrule[0.5pt]
    Ion. &  9,044 &  181,225  & 0.008 & 6.18 \\
    NSL &  142,856   &  85,435,573  &0.020 & 16.28\\
    M.T. &  303,025 &  2,240,917 & 0.015 & 9.70\\
    CPU &  206,337  & 4,601,219 & 0.015 & 9.22 \\
    INSECTS-Abr & 520,843 & 24,662,015 & 0.017 & 38.08\\
    \bottomrule[1.5pt]
        \end{tabular}
    }
\end{table}

\noindent
\blue{\textbf{Reccurring Patterns:} We partition INSECTS-Abr dataset into five sequential and equal subsets, denoted as $p_1$ to $p_5$. Then, we further split each subset into one training set and one test set, and derive $p_{1-train}/p_{1-test}$ to $p_{5-train}/p_{5-test}$ correspondingly.
Using these subsets, we conduct two groups of experiments. 
In Group I, we train \name on $p_{1-train}$ and then fine-tune \name on $p_{2-train}$ to $p_{5-train}$ respectively, and report the detection performance of \name. As for the experiments of Group II, we only train \name on $p_{1-train}$ once, and then report the trained \name on all test sets, namely $p_{1-test}$ to $p_{5-test}$.
Results summarized in Table~\ref{tab:reccurring} show that (1) the fine-tuning strategy is not effective in OAD, which may lead to worse performance than the model initially trained but without fine-tuning, e.g., the Group I model archives an AUCROC of 0.702 on $p_{4-test}$, which is much worse than 0.754 of the Group II model without further fine-tuning.
(2) The model of Group II archives an AUCROC of 0.768, which is only slightly worse than 0.804 obtained on $p_{1-test}$, and obtains an even higher AUCROC of 0.837 on $p_{5-test}$.
These two findings suggest that the majority of anomaly patterns are indeed already encompassed on the status historical data, namely $p_{1-train}$, and the model of Group II only trained on $p_{1-train}$ can detect most anomalies on the subsequent unseen test sets, which achieve so by learning the recurring central concepts encompassed in the static historical data.}

\subsection{\blue{Efficiency}}
\label{sec:exp_efficiency}

\blue{We evaluate the training and inference efficiency of \name by quantifying its throughput on benchmark datasets of varying sizes and dimensions. Throughput, denoted as the number of processed samples per second, is evaluated for both training and inference.
Results in Table \ref{tab:efficiency} show that \name is rather efficient in both training and inference on datasets of various sizes and dimensions.
This indicates that \name exhibits high computational efficiency
and thus supports rapid response rates for real-time OAD applications.
}
\blue{Another issue worth concerning is the training time and the maximum memory usage (peak memory). Considering \name's adaptive offline update capability to handle concept drift, achieving a shorter training time and a smaller peak memory becomes especially desirable.
Table \ref{tab:efficiency} reports the average training time for each epoch and the peak memory. 
The results demonstrate that \name requires negligible training time and takes low memory usage across these datasets.
This is mainly due to the lightweight design of the key modules of \name, as discussed in detail in Section~\ref{sec:analysis_and_discussion}.
To provide further insights into this matter, we conduct tests to assess the efficiency impact of the IEC and DSD modules. Specifically, we compare peak memory usage and average training time per epoch on CPU of \name with and without IEC and DSD.
Results show that the introduction of the IEC and DSD modules incurs a negligible increase in training time (0.014s and 0.007s, respectively) and peak memory usage (5.17MB and 2.61MB, respectively). 
Further, our ablation studies in Table \ref{tab:ablation} demonstrate that integrating these two modules into \name enhances the performance by a large margin.
These findings corroborate the efficiency and effectiveness of the IEC and DSD in our framework, which is well-suited for real-time applications and supports high-performance OAD with high efficiency.
}

\begin{figure}[t]
    \centering
    \includegraphics[width=0.9\linewidth]{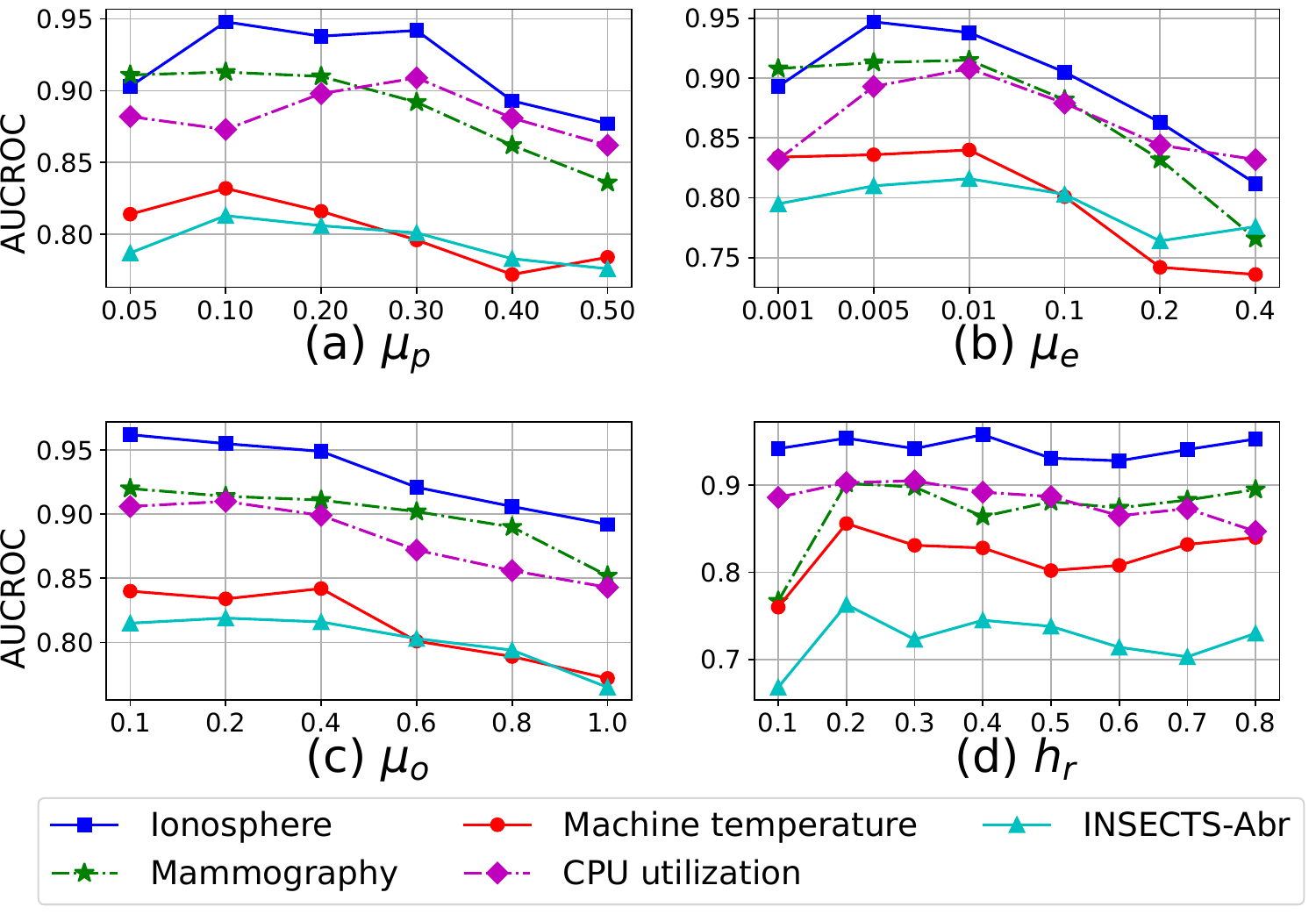}\vspace{-2mm}
    \caption{\blue{ Sensitivity test of parameters $\mu_{p}$, $\mu_{e}$, $\mu_{o}$, and $\Delta L$.
    }}
    \label{fig:sentitive}
 \vspace{-2mm}
\end{figure}
\begin{table}[t!]
    \small
    \centering
    \renewcommand{\arraystretch}{0.85}
    \caption{ \blue{The performance and inference time under different $\Delta L$.}}\vspace{-2mm}
    \label{tab:sentitive}
    \resizebox{1.0\columnwidth}{!}{
        \begin{tabular}{ c c c c c c c }
    \toprule[1.5pt]
   	Window Size $\Delta L$ & 32  &  48  & 64 & 96 & 128 & 256 \\ 
    \hline	
        AUCROC  & 0.875 & 0.862 & 0.856 & 0.832 & 0.805  & 0.798 \\
	Time(s) & 253 & 248 & 188 & 173 & 133 & 98 \\
    \bottomrule[1.5pt]
    \end{tabular}
    }
    \vspace{-2mm}
\end{table}
\subsection{Sensitivity Study}
\label{sec:exp_sensitivity}

\blue{First, we focus on the three critical threshold hyperparameters: $\mu_{p}$, $\mu_{e}$, and $\mu_{o}$.}
These hyperparameters control dynamic concept adaptation and the frequency of model updates.
Adopting low thresholds for $\mu_{p}$ and $\mu_{e}$ may result in an excessive number of samples being allocated to the Dynamic Shift Detector (DSD), even when the current concept remains unchanged.
Conversely, excessively high thresholds could hinder the detection of concept drift when it arises.
Experimental results, as depicted in Figure~\ref{fig:sentitive}, suggest that the optimal settings for $\mu_{p}$ lie between 0.1 and 0.2, and for $\mu_{e}$ between 0.005 and 0.01.
In the absence of prior knowledge about a dataset, we recommend these default parameters as initial values, followed by a grid search to find better hyperparameters.
As for $\mu_{o}$, this hyperparameter denotes the number of samples that exceed the update threshold within the sliding window.
To evaluate the effect of the offline updating strategy, we scrutinized the ratio of $\mu_{o}$ to the window size.
Our results indicate that a smaller $\mu_{o}$ value improves performance.
However, a small $\mu_{o}$ can lead to frequent offline updates, and thus reduce model efficiency.
Therefore, a balance between performance and efficiency can be attained by setting $\mu_{o}$ between 0.1 and 0.4, as shown in Figure~\ref{fig:sentitive}.

\blue{Next, we evaluate the historical data ratio $h_r$ within the range of 0.1 to 0.8.
One key observation is that when $h_r$ is too small, the performance notably declines due to the inadequacy of historical data. This inadequacy results in a failure to acquire adequately informative central concepts.
As the ratio increases, there is a noticeable improvement in performance.
However, a higher ratio does not consistently guarantee better performance. For instance, in the CPU and INSECTS-Abr datasets, increasing the ratio actually leads to a decline in performance. This observation suggests that the model might become more susceptible to overfitting the training data, thus causing a reduction in performance during inference.
While the targeted OUS is more efficient and performs better than training models directly on more data. 
}

\blue{Furthermore, we evaluate the impact of the window size $\Delta L$ on both performance and efficiency. This parameter is central for OUS.
The results in Table~\ref{tab:sentitive} show that an excessively small $\Delta L$ results in frequent model updates, which can somewhat ensure a certain level of AUCROC performance but at the expense of efficiency. Conversely, an overly large $\Delta L$ reduces inference time but compromises the model's capacity to timely capture concept changes, resulting in performance degradation.
Fortunately, as discussed in Section~\ref{sec:exp_efficiency}, the update time of the model is minimal, and it maintains commendable performance across a wide range of $\Delta L$.
This allows users to tailor the window size to their specific requirements in practical applications.
}


\subsection{Interpretability}
\label{sec:exp_interpretability}
\begin{figure}[t]
    \centering\vspace{-3mm}
    \includegraphics[width=0.9\linewidth]{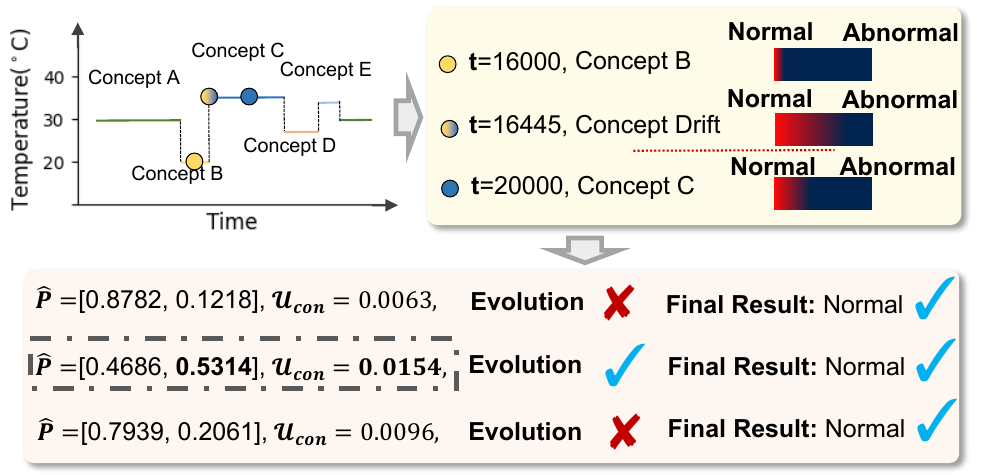}\vspace{-2mm}
    \caption{Interpretability with respect to model evolution.
    }
    \label{fig:interpretability}
    \vspace{-4mm}
\end{figure}

\blue{
To provide a clearer understanding of the uncertainty modeled by IEC, we offer an interpretation from a semantic perspective.
High uncertainty in anomaly detection often arises when the model confronts unfamiliar data concepts.
Figure~\ref{fig:interpretability} serves as an illustrative example of how \name generates interpretable results for different time steps and concepts, and how it accurately identifies and rapidly adapts to concept drift based on probability distribution and concept uncertainty as observed on the real-world dataset INSECTS.}
We identify three representative data points from two different concepts and their concept drift point for illustration.
For the first data point, \name generates a probability distribution characterized by small entropy and low uncertainty, enabling the model to correctly classify it as a normal sample.
For the second data point, \name produces a probability distribution with high concept uncertainty, indicating the presence of concept drift.
In this case, \name transitions into the dynamic mode via the IEC, and the DSD learns the parameter drift of the base detection model and accurately classifies it as a normal sample.
By examining the output of the third point, we can notice that the model effectively adapts to the new concept, indicated by the prediction with low concept uncertainty.
These findings validate the capability of \name to provide more interpretable and trustworthy detection for better user understanding.



\begin{figure}[t]
    \centering
    \vspace{-3mm}
    \includegraphics[width=1\linewidth]{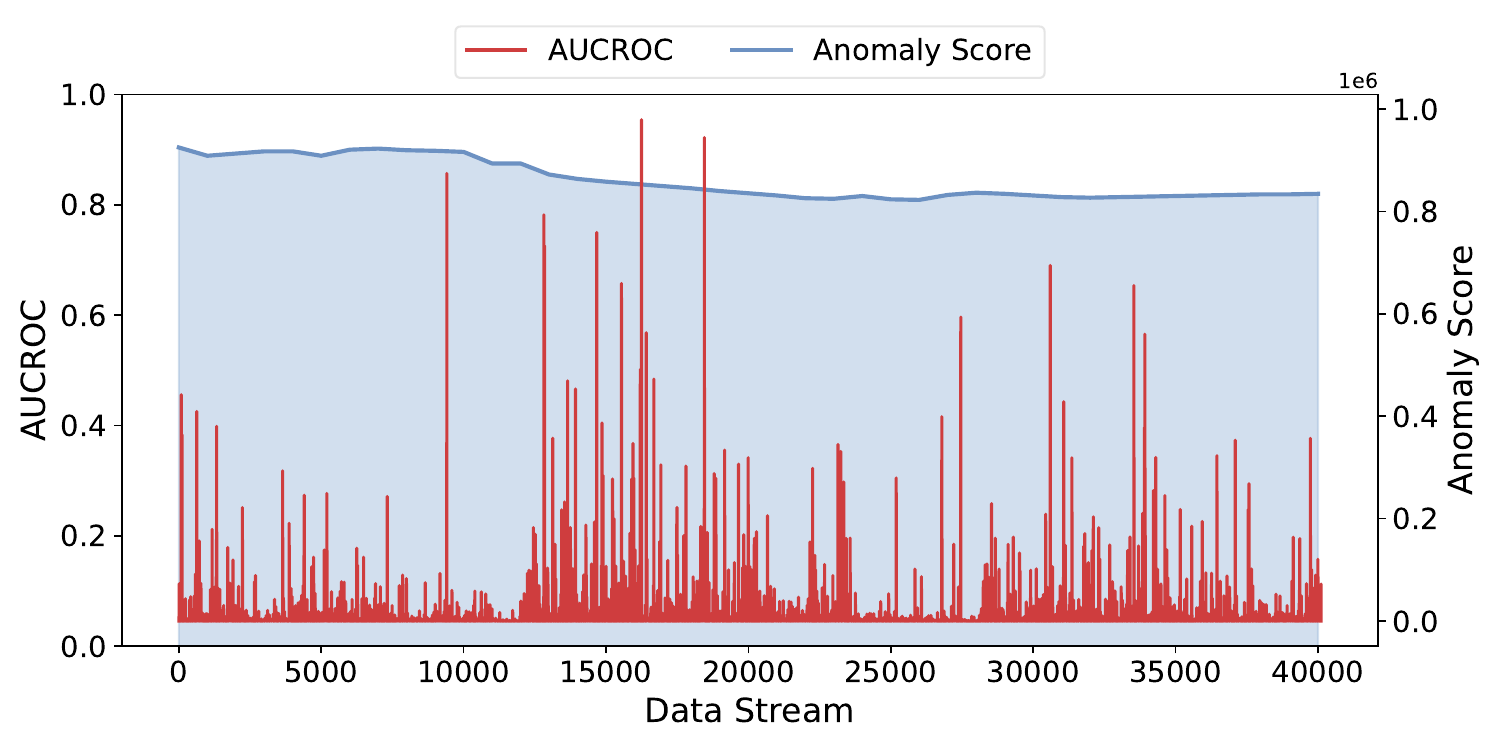}\vspace{-3mm}
    \caption{\blue{Timely output results of \name on Flink.} }
    \label{fig:flink}
    \vspace{-4mm}
\end{figure}

\subsection{\blue{Integration on Flink}}
\label{sec:flink}
\blue{
Stream processing engines play a pivotal role in deploying
OAD frameworks.
To illustrate how \name can function as a component of a larger-scale system, 
we integrate \name into Apache Flink~\cite{carbone2015apache}, a framework and distributed processing engine for stateful computations over unbounded and bounded data streams. 
Flink wraps \name inside a Flink operator, where Flink helps to establish the required environment, manage resources, read/write the data with versatile connectors, and handle failures.
Building streaming workloads upon Flink empowers \name with real-time data processing pipelines, large-scale exploratory data analysis, and ETL processes.
The experiments are conducted on the INSECT-Abr dataset.
The timely output results, as illustrated in Figure~\ref{fig:flink}, validate a seamless integration of \name with a stream processing engine, confirming its efficacy in supporting real-time anomaly detection.
}
\section{Related Work}
\label{sec:related work}

\noindent
\revise{\highlight{Anomaly Detection.}}
Anomaly detection (AD) has been extensively studied in various fields such as computer networks \cite{tartakovsky2012efficient, nawir2019effective}, intrusion detection \cite{huang2015streaming,falcao2019quantitative}, healthcare \cite{vsabic2021healthcare, wang2020guardhealth}, and finance \cite{hilal2022financial}. 
Traditional approaches for AD include statistical methods \cite{breunig2000lof}, clustering algorithms \cite{guo2003knn, angiulli2007detecting}, classification-based techniques \cite{liu2008isolation}, and nearest neighbor-based methods \cite{tian2014anomaly,boniol2021unsupervised}.
With the advent of deep learning, autoencoder-based techniques have become popular for AD \cite{an2015variational}. Autoencoders learn to reconstruct input data by minimizing the reconstruction error, where a higher error is an indication of an anomaly. 
\cite{sakurada2014anomaly} proposes a stacked autoencoder for detecting anomalies in credit card transactions. The method achieved good performance but suffered from high false positive rates. To address this issue, \cite{makhzani2015winner} introduces a sparse autoencoder, which achieves better performance in terms of false positive rates. 
There are also some extensions for traditional autoencoders on AD. ADAE \cite{vu2019anomaly} proposes an Adversarial Dual Autoencoder framework for AD, which uses two autoencoders in a dual structure to improve the representation power of the model. \cite{li2020anomaly} presents a smoothness-inducing sequential variational auto-encoder (VAE) model for the robust estimation and AD of multidimensional time series.
However, the above approaches have not considered processing data in a streaming fashion and usually require large amounts of training data in an offline setting, and thus cannot be applied to online AD.




\vspace{1mm}
\noindent
\highlight{Online Anomaly Detection.}
\blue{Online anomaly detection (OAD) aims to promptly identify abnormal behavior or events in real-time data streams.
Notably, compared to AD, OAD faces the challenge of concept drift.
Previous works \cite{bifet2007learning,cavalcante2016fedd,huang2014detecting} rely on sliding windows and primarily focus on detecting alterations in specific statistical characteristics (e.g., mean values) of streaming data or its individual features to identify concept drift.
However, these approaches prove insufficient for OAD, since concept drift within OAD can encompass more intricate changes in data distribution, including shifts, correlations, and transformations of data patterns, extending beyond the scope of basic statistical features like mean values.}


\blue{
Different anomaly detection approaches span various paradigms, such as isolation forest~\cite{liu2008isolation} (IF), k nearest neighbors (kNN), local outlier factor~\cite{breunig2000lof} (LOF), and deep neural networks (DNNs).
These approaches have been extensively adapted for OAD tasks, leading to the rise of two prevailing OAD approaches: incremental learning-based \cite{guha2016robust, na2018dilof, bhatia2021mstream, bhatia2022memstream,boniol2021sand} and ensemble-based approaches \cite{tan2011fast, ding2013anomaly, sathe2016subspace, pevny2016loda,manzoor2018xstream,yoon2022adaptive,mirsky2018kitsune,gopalan2019pidforest}.
}

For incremental learning-based approaches, MStream \cite{bhatia2021mstream} was proposed as a locality-sensitive hash functions-based method to employ two types of hash functions and utilizes dimensionality reduction techniques based on autoencoder. MemStream \cite{bhatia2022memstream} utilizes a denoising autoencoder to acquire representations and a memory module to capture the evolving trends in data without requiring labeled data. \revise{Moreover, there are several incremental algorithms that focus on computing similarity measures, such as distance metrics, to address the challenge of identifying anomalies in evolving sequences~\cite{boniol2021sand,tran2020real,lu2022matrix}.
}

The ensemble-based methods have also gained popularity in recent years. iForestASD \cite{ding2013anomaly} combines sliding windows with Isolation Forest for the purpose of detecting anomalies in streaming data. xStream \cite{manzoor2018xstream} uses lower-dimensional projections of the data points, which are seamlessly updated to accommodate newly emerging features, and performs outlier detection through an ensemble of randomized partitions. It handles temporal shifts by a window-based approach where bin counts accumulated in the previous window are used to score points in the current window. ARCUS \cite{yoon2022adaptive} proposes adaptive model pooling to manage multiple models in both inference and adaptation to handle evolving data streams with multiple and time-varying concept drifts, achieving versatile anomaly detection performance.


\blue{However, these methods suffer from limitations in their ability to adapt to evolving data streams, since they need to update their detection models to accommodate the changing concepts, typically by retraining, fine-tuning, or updating certain model statuses.
Moreover, the effectiveness of these techniques is inherently bounded by their quantity, such as the number of pre-trained models in ensemble-based methods, which necessitates a trade-off between efficiency and performance.
Our work aims to address these challenges by proposing a novel framework that dynamically adapts to concept drift without the need for additional fine-tuning or retraining, which guarantees both effectiveness and efficiency while providing interpretable decision-making.}




\section{Conclusions}
\label{sec:conclusion}

In this paper, we present a novel framework \name for online anomaly detection (OAD), which addresses the challenge of concept drift in an effective, efficient, and interpretable manner. By leveraging a static concept-aware detector trained on historical data, \name captures and handles recurring central concepts, while dynamically adapting to new concepts in evolving data streams using a lightweight drift detection controller and a hypernetwork-based parameter shift technique. 
The evidential deep learning-based drift detection controller enables efficient and interpretable concept drift detection.
Our experimental study demonstrates that \name outperforms existing OAD approaches in various scenarios and facilitates valuable interpretability. 


\newpage

\balance

\bibliographystyle{ACM-Reference-Format}
\bibliography{reference}


\begin{thebibliography}{93}


\ifx \showCODEN    \undefined \def \showCODEN     #1{\unskip}     \fi
\ifx \showDOI      \undefined \def \showDOI       #1{#1}\fi
\ifx \showISBNx    \undefined \def \showISBNx     #1{\unskip}     \fi
\ifx \showISBNxiii \undefined \def \showISBNxiii  #1{\unskip}     \fi
\ifx \showISSN     \undefined \def \showISSN      #1{\unskip}     \fi
\ifx \showLCCN     \undefined \def \showLCCN      #1{\unskip}     \fi
\ifx \shownote     \undefined \def \shownote      #1{#1}          \fi
\ifx \showarticletitle \undefined \def \showarticletitle #1{#1}   \fi
\ifx \showURL      \undefined \def \showURL       {\relax}        \fi
\providecommand\bibfield[2]{#2}
\providecommand\bibinfo[2]{#2}
\providecommand\natexlab[1]{#1}
\providecommand\showeprint[2][]{arXiv:#2}

\bibitem[\protect\citeauthoryear{??}{kdd}{1999}]%
        {kdd99-web}
 \bibinfo{year}{1999}\natexlab{}.
\newblock \bibinfo{title}{KDD Cup Dataset}.
\newblock \bibinfo{howpublished}{\url{http://kdd.ics.uci.edu/databases/kddcup99/kddcup99.html}. Accessed:2023-07}.
\newblock


\bibitem[\protect\citeauthoryear{Abdulaal, Liu, and Lancewicki}{Abdulaal et~al\mbox{.}}{2021}]%
        {abdulaal2021practical}
\bibfield{author}{\bibinfo{person}{Ahmed Abdulaal}, \bibinfo{person}{Zhuanghua Liu}, {and} \bibinfo{person}{Tomer Lancewicki}.} \bibinfo{year}{2021}\natexlab{}.
\newblock \showarticletitle{Practical approach to asynchronous multivariate time series anomaly detection and localization}. In \bibinfo{booktitle}{\emph{Proceedings of the 27th ACM SIGKDD conference on knowledge discovery \& data mining}}. \bibinfo{pages}{2485--2494}.
\newblock


\bibitem[\protect\citeauthoryear{Ahmad, Lavin, Purdy, and Agha}{Ahmad et~al\mbox{.}}{2017}]%
        {ahmad2017unsupervised}
\bibfield{author}{\bibinfo{person}{Subutai Ahmad}, \bibinfo{person}{Alexander Lavin}, \bibinfo{person}{Scott Purdy}, {and} \bibinfo{person}{Zuha Agha}.} \bibinfo{year}{2017}\natexlab{}.
\newblock \showarticletitle{Unsupervised real-time anomaly detection for streaming data}.
\newblock \bibinfo{journal}{\emph{Neurocomputing}}  \bibinfo{volume}{262} (\bibinfo{year}{2017}), \bibinfo{pages}{134--147}.
\newblock


\bibitem[\protect\citeauthoryear{An and Cho}{An and Cho}{2015}]%
        {an2015variational}
\bibfield{author}{\bibinfo{person}{Jinwon An} {and} \bibinfo{person}{Sungzoon Cho}.} \bibinfo{year}{2015}\natexlab{}.
\newblock \showarticletitle{Variational autoencoder based anomaly detection using reconstruction probability}.
\newblock \bibinfo{journal}{\emph{Special lecture on IE}} \bibinfo{volume}{2}, \bibinfo{number}{1} (\bibinfo{year}{2015}), \bibinfo{pages}{1--18}.
\newblock


\bibitem[\protect\citeauthoryear{Angiulli and Fassetti}{Angiulli and Fassetti}{2007}]%
        {angiulli2007detecting}
\bibfield{author}{\bibinfo{person}{Fabrizio Angiulli} {and} \bibinfo{person}{Fabio Fassetti}.} \bibinfo{year}{2007}\natexlab{}.
\newblock \showarticletitle{Detecting distance-based outliers in streams of data}. In \bibinfo{booktitle}{\emph{Proceedings of the sixteenth ACM conference on Conference on information and knowledge management}}. \bibinfo{pages}{811--820}.
\newblock


\bibitem[\protect\citeauthoryear{Audibert, Michiardi, Guyard, Marti, and Zuluaga}{Audibert et~al\mbox{.}}{2020}]%
        {audibert2020usad}
\bibfield{author}{\bibinfo{person}{Julien Audibert}, \bibinfo{person}{Pietro Michiardi}, \bibinfo{person}{Fr{\'e}d{\'e}ric Guyard}, \bibinfo{person}{S{\'e}bastien Marti}, {and} \bibinfo{person}{Maria~A Zuluaga}.} \bibinfo{year}{2020}\natexlab{}.
\newblock \showarticletitle{Usad: Unsupervised anomaly detection on multivariate time series}. In \bibinfo{booktitle}{\emph{Proceedings of the 26th ACM SIGKDD International Conference on Knowledge Discovery \& Data Mining}}. \bibinfo{pages}{3395--3404}.
\newblock


\bibitem[\protect\citeauthoryear{Bhatia, Jain, Li, Kumar, and Hooi}{Bhatia et~al\mbox{.}}{2021}]%
        {bhatia2021mstream}
\bibfield{author}{\bibinfo{person}{Siddharth Bhatia}, \bibinfo{person}{Arjit Jain}, \bibinfo{person}{Pan Li}, \bibinfo{person}{Ritesh Kumar}, {and} \bibinfo{person}{Bryan Hooi}.} \bibinfo{year}{2021}\natexlab{}.
\newblock \showarticletitle{Mstream: Fast anomaly detection in multi-aspect streams}. In \bibinfo{booktitle}{\emph{Proceedings of the Web Conference 2021}}. \bibinfo{pages}{3371--3382}.
\newblock


\bibitem[\protect\citeauthoryear{Bhatia, Jain, Srivastava, Kawaguchi, and Hooi}{Bhatia et~al\mbox{.}}{2022}]%
        {bhatia2022memstream}
\bibfield{author}{\bibinfo{person}{Siddharth Bhatia}, \bibinfo{person}{Arjit Jain}, \bibinfo{person}{Shivin Srivastava}, \bibinfo{person}{Kenji Kawaguchi}, {and} \bibinfo{person}{Bryan Hooi}.} \bibinfo{year}{2022}\natexlab{}.
\newblock \showarticletitle{MemStream: Memory-Based Streaming Anomaly Detection}. In \bibinfo{booktitle}{\emph{Proceedings of the ACM Web Conference 2022}}. \bibinfo{pages}{610--621}.
\newblock


\bibitem[\protect\citeauthoryear{Bifet and Gavalda}{Bifet and Gavalda}{2007}]%
        {bifet2007learning}
\bibfield{author}{\bibinfo{person}{Albert Bifet} {and} \bibinfo{person}{Ricard Gavalda}.} \bibinfo{year}{2007}\natexlab{}.
\newblock \showarticletitle{Learning from time-changing data with adaptive windowing}. In \bibinfo{booktitle}{\emph{Proceedings of the 2007 SIAM international conference on data mining}}. SIAM, \bibinfo{pages}{443--448}.
\newblock


\bibitem[\protect\citeauthoryear{Boniol, Linardi, Roncallo, Palpanas, Meftah, and Remy}{Boniol et~al\mbox{.}}{2021a}]%
        {boniol2021unsupervised}
\bibfield{author}{\bibinfo{person}{Paul Boniol}, \bibinfo{person}{Michele Linardi}, \bibinfo{person}{Federico Roncallo}, \bibinfo{person}{Themis Palpanas}, \bibinfo{person}{Mohammed Meftah}, {and} \bibinfo{person}{Emmanuel Remy}.} \bibinfo{year}{2021}\natexlab{a}.
\newblock \showarticletitle{Unsupervised and scalable subsequence anomaly detection in large data series}.
\newblock \bibinfo{journal}{\emph{The VLDB Journal}} (\bibinfo{year}{2021}), \bibinfo{pages}{1--23}.
\newblock


\bibitem[\protect\citeauthoryear{Boniol, Paparrizos, Palpanas, and Franklin}{Boniol et~al\mbox{.}}{2021b}]%
        {boniol2021sand}
\bibfield{author}{\bibinfo{person}{Paul Boniol}, \bibinfo{person}{John Paparrizos}, \bibinfo{person}{Themis Palpanas}, {and} \bibinfo{person}{Michael~J Franklin}.} \bibinfo{year}{2021}\natexlab{b}.
\newblock \showarticletitle{SAND: streaming subsequence anomaly detection}.
\newblock \bibinfo{journal}{\emph{Proceedings of the VLDB Endowment}} \bibinfo{volume}{14}, \bibinfo{number}{10} (\bibinfo{year}{2021}), \bibinfo{pages}{1717--1729}.
\newblock


\bibitem[\protect\citeauthoryear{Breunig, Kriegel, Ng, and Sander}{Breunig et~al\mbox{.}}{2000}]%
        {breunig2000lof}
\bibfield{author}{\bibinfo{person}{Markus~M Breunig}, \bibinfo{person}{Hans-Peter Kriegel}, \bibinfo{person}{Raymond~T Ng}, {and} \bibinfo{person}{J{\"o}rg Sander}.} \bibinfo{year}{2000}\natexlab{}.
\newblock \showarticletitle{LOF: identifying density-based local outliers}. In \bibinfo{booktitle}{\emph{Proceedings of the 2000 ACM SIGMOD international conference on Management of data}}. \bibinfo{pages}{93--104}.
\newblock


\bibitem[\protect\citeauthoryear{Cai, Zheng, Chen, Jagadish, Ooi, and Zhang}{Cai et~al\mbox{.}}{2021}]%
        {cai2021arm}
\bibfield{author}{\bibinfo{person}{Shaofeng Cai}, \bibinfo{person}{Kaiping Zheng}, \bibinfo{person}{Gang Chen}, \bibinfo{person}{HV Jagadish}, \bibinfo{person}{Beng~Chin Ooi}, {and} \bibinfo{person}{Meihui Zhang}.} \bibinfo{year}{2021}\natexlab{}.
\newblock \showarticletitle{Arm-Net: Adaptive relation modeling network for structured data}. In \bibinfo{booktitle}{\emph{Proceedings of the 2021 International Conference on Management of Data}}. \bibinfo{pages}{207--220}.
\newblock


\bibitem[\protect\citeauthoryear{Carbone, Katsifodimos, Ewen, Markl, Haridi, and Tzoumas}{Carbone et~al\mbox{.}}{2015}]%
        {carbone2015apache}
\bibfield{author}{\bibinfo{person}{Paris Carbone}, \bibinfo{person}{Asterios Katsifodimos}, \bibinfo{person}{Stephan Ewen}, \bibinfo{person}{Volker Markl}, \bibinfo{person}{Seif Haridi}, {and} \bibinfo{person}{Kostas Tzoumas}.} \bibinfo{year}{2015}\natexlab{}.
\newblock \showarticletitle{Apache flink: Stream and batch processing in a single engine}.
\newblock \bibinfo{journal}{\emph{The Bulletin of the Technical Committee on Data Engineering}} \bibinfo{volume}{38}, \bibinfo{number}{4} (\bibinfo{year}{2015}).
\newblock


\bibitem[\protect\citeauthoryear{Cavalcante, Minku, and Oliveira}{Cavalcante et~al\mbox{.}}{2016}]%
        {cavalcante2016fedd}
\bibfield{author}{\bibinfo{person}{Rodolfo~C Cavalcante}, \bibinfo{person}{Leandro~L Minku}, {and} \bibinfo{person}{Adriano~LI Oliveira}.} \bibinfo{year}{2016}\natexlab{}.
\newblock \showarticletitle{Fedd: Feature extraction for explicit concept drift detection in time series}. In \bibinfo{booktitle}{\emph{2016 International Joint Conference on Neural Networks (IJCNN)}}. IEEE, \bibinfo{pages}{740--747}.
\newblock


\bibitem[\protect\citeauthoryear{Chandola, Banerjee, and Kumar}{Chandola et~al\mbox{.}}{2009}]%
        {chandola2009anomaly}
\bibfield{author}{\bibinfo{person}{Varun Chandola}, \bibinfo{person}{Arindam Banerjee}, {and} \bibinfo{person}{Vipin Kumar}.} \bibinfo{year}{2009}\natexlab{}.
\newblock \showarticletitle{Anomaly detection: A survey}.
\newblock \bibinfo{journal}{\emph{ACM computing surveys (CSUR)}} \bibinfo{volume}{41}, \bibinfo{number}{3} (\bibinfo{year}{2009}), \bibinfo{pages}{1--58}.
\newblock


\bibitem[\protect\citeauthoryear{Chawathe and Garcia-Molina}{Chawathe and Garcia-Molina}{1997}]%
        {chawathe1997meaningful}
\bibfield{author}{\bibinfo{person}{Sudarshan~S Chawathe} {and} \bibinfo{person}{Hector Garcia-Molina}.} \bibinfo{year}{1997}\natexlab{}.
\newblock \showarticletitle{Meaningful change detection in structured data}.
\newblock \bibinfo{journal}{\emph{ACM SIGMOD Record}} \bibinfo{volume}{26}, \bibinfo{number}{2} (\bibinfo{year}{1997}), \bibinfo{pages}{26--37}.
\newblock


\bibitem[\protect\citeauthoryear{Chen, Liu, Su, Zhang, Ling, Yang, and Lyu}{Chen et~al\mbox{.}}{2022}]%
        {chen2022adaptive}
\bibfield{author}{\bibinfo{person}{Zhuangbin Chen}, \bibinfo{person}{Jinyang Liu}, \bibinfo{person}{Yuxin Su}, \bibinfo{person}{Hongyu Zhang}, \bibinfo{person}{Xiao Ling}, \bibinfo{person}{Yongqiang Yang}, {and} \bibinfo{person}{Michael~R Lyu}.} \bibinfo{year}{2022}\natexlab{}.
\newblock \showarticletitle{Adaptive performance anomaly detection for online service systems via pattern sketching}. In \bibinfo{booktitle}{\emph{Proceedings of the 44th International Conference on Software Engineering}}. \bibinfo{pages}{61--72}.
\newblock


\bibitem[\protect\citeauthoryear{Dau, Keogh, Kamgar, Yeh, Zhu, Gharghabi, Ratanamahatana, Yanping, Hu, Begum, Bagnall, Mueen, Batista, and Hexagon-ML}{Dau et~al\mbox{.}}{2021}]%
        {UCRArchive2021}
\bibfield{author}{\bibinfo{person}{Hoang~Anh Dau}, \bibinfo{person}{Eamonn Keogh}, \bibinfo{person}{Kaveh Kamgar}, \bibinfo{person}{Chin-Chia~Michael Yeh}, \bibinfo{person}{Yan Zhu}, \bibinfo{person}{Shaghayegh Gharghabi}, \bibinfo{person}{Chotirat~Ann Ratanamahatana}, \bibinfo{person}{Yanping}, \bibinfo{person}{Bing Hu}, \bibinfo{person}{Nurjahan Begum}, \bibinfo{person}{Anthony Bagnall}, \bibinfo{person}{Abdullah Mueen}, \bibinfo{person}{Gustavo Batista}, {and} \bibinfo{person}{Hexagon-ML}.} \bibinfo{year}{2021}\natexlab{}.
\newblock \bibinfo{title}{The UCR Time Series Classification Archive}.
\newblock
\newblock
\newblock
\shownote{\url{https://www.cs.ucr.edu/~eamonn/time_series_data_2018/UCR_TimeSeriesAnomalyDatasets2021.zip}. Accessed:2023-07.}


\bibitem[\protect\citeauthoryear{Deng and Li}{Deng and Li}{2022}]%
        {deng2022anomaly}
\bibfield{author}{\bibinfo{person}{Hanqiu Deng} {and} \bibinfo{person}{Xingyu Li}.} \bibinfo{year}{2022}\natexlab{}.
\newblock \showarticletitle{Anomaly detection via reverse distillation from one-class embedding}. In \bibinfo{booktitle}{\emph{Proceedings of the IEEE/CVF Conference on Computer Vision and Pattern Recognition}}. \bibinfo{pages}{9737--9746}.
\newblock


\bibitem[\protect\citeauthoryear{Ding and Fei}{Ding and Fei}{2013}]%
        {ding2013anomaly}
\bibfield{author}{\bibinfo{person}{Zhiguo Ding} {and} \bibinfo{person}{Minrui Fei}.} \bibinfo{year}{2013}\natexlab{}.
\newblock \showarticletitle{An anomaly detection approach based on isolation forest algorithm for streaming data using sliding window}.
\newblock \bibinfo{journal}{\emph{IFAC Proceedings Volumes}} \bibinfo{volume}{46}, \bibinfo{number}{20} (\bibinfo{year}{2013}), \bibinfo{pages}{12--17}.
\newblock


\bibitem[\protect\citeauthoryear{Falc{\~a}o, Zoppi, Silva, Santos, Fonseca, Ceccarelli, and Bondavalli}{Falc{\~a}o et~al\mbox{.}}{2019}]%
        {falcao2019quantitative}
\bibfield{author}{\bibinfo{person}{Filipe Falc{\~a}o}, \bibinfo{person}{Tommaso Zoppi}, \bibinfo{person}{Caio Barbosa~Viera Silva}, \bibinfo{person}{Anderson Santos}, \bibinfo{person}{Baldoino Fonseca}, \bibinfo{person}{Andrea Ceccarelli}, {and} \bibinfo{person}{Andrea Bondavalli}.} \bibinfo{year}{2019}\natexlab{}.
\newblock \showarticletitle{Quantitative comparison of unsupervised anomaly detection algorithms for intrusion detection}. In \bibinfo{booktitle}{\emph{Proceedings of the 34th ACM/SIGAPP Symposium on Applied Computing}}. \bibinfo{pages}{318--327}.
\newblock


\bibitem[\protect\citeauthoryear{Gao, Liu, Wu, Shi, Shi, and Zhuang}{Gao et~al\mbox{.}}{2023}]%
        {gao2023reliable}
\bibfield{author}{\bibinfo{person}{Zheyao Gao}, \bibinfo{person}{Yuanye Liu}, \bibinfo{person}{Fuping Wu}, \bibinfo{person}{NanNan Shi}, \bibinfo{person}{Yuxin Shi}, {and} \bibinfo{person}{Xiahai Zhuang}.} \bibinfo{year}{2023}\natexlab{}.
\newblock \showarticletitle{A Reliable and Interpretable Framework of Multi-view Learning for Liver Fibrosis Staging}.
\newblock \bibinfo{journal}{\emph{arXiv preprint arXiv:2306.12054}} (\bibinfo{year}{2023}).
\newblock


\bibitem[\protect\citeauthoryear{Gong, Liu, Le, Saha, Mansour, Venkatesh, and Hengel}{Gong et~al\mbox{.}}{2019}]%
        {gong2019memorizing}
\bibfield{author}{\bibinfo{person}{Dong Gong}, \bibinfo{person}{Lingqiao Liu}, \bibinfo{person}{Vuong Le}, \bibinfo{person}{Budhaditya Saha}, \bibinfo{person}{Moussa~Reda Mansour}, \bibinfo{person}{Svetha Venkatesh}, {and} \bibinfo{person}{Anton van~den Hengel}.} \bibinfo{year}{2019}\natexlab{}.
\newblock \showarticletitle{Memorizing normality to detect anomaly: Memory-augmented deep autoencoder for unsupervised anomaly detection}. In \bibinfo{booktitle}{\emph{Proceedings of the IEEE/CVF International Conference on Computer Vision}}. \bibinfo{pages}{1705--1714}.
\newblock


\bibitem[\protect\citeauthoryear{Gopalan, Sharan, and Wieder}{Gopalan et~al\mbox{.}}{2019}]%
        {gopalan2019pidforest}
\bibfield{author}{\bibinfo{person}{Parikshit Gopalan}, \bibinfo{person}{Vatsal Sharan}, {and} \bibinfo{person}{Udi Wieder}.} \bibinfo{year}{2019}\natexlab{}.
\newblock \showarticletitle{Pidforest: anomaly detection via partial identification}.
\newblock \bibinfo{journal}{\emph{Advances in Neural Information Processing Systems}}  \bibinfo{volume}{32} (\bibinfo{year}{2019}).
\newblock


\bibitem[\protect\citeauthoryear{Guha, Mishra, Roy, and Schrijvers}{Guha et~al\mbox{.}}{2016}]%
        {guha2016robust}
\bibfield{author}{\bibinfo{person}{Sudipto Guha}, \bibinfo{person}{Nina Mishra}, \bibinfo{person}{Gourav Roy}, {and} \bibinfo{person}{Okke Schrijvers}.} \bibinfo{year}{2016}\natexlab{}.
\newblock \showarticletitle{Robust random cut forest based anomaly detection on streams}. In \bibinfo{booktitle}{\emph{International conference on machine learning}}. PMLR, \bibinfo{pages}{2712--2721}.
\newblock


\bibitem[\protect\citeauthoryear{Guo, Wang, Bell, Bi, and Greer}{Guo et~al\mbox{.}}{2003}]%
        {guo2003knn}
\bibfield{author}{\bibinfo{person}{Gongde Guo}, \bibinfo{person}{Hui Wang}, \bibinfo{person}{David Bell}, \bibinfo{person}{Yaxin Bi}, {and} \bibinfo{person}{Kieran Greer}.} \bibinfo{year}{2003}\natexlab{}.
\newblock \showarticletitle{KNN model-based approach in classification}. In \bibinfo{booktitle}{\emph{On The Move to Meaningful Internet Systems 2003: CoopIS, DOA, and ODBASE: OTM Confederated International Conferences, CoopIS, DOA, and ODBASE 2003, Catania, Sicily, Italy, November 3-7, 2003. Proceedings}}. Springer, \bibinfo{pages}{986--996}.
\newblock


\bibitem[\protect\citeauthoryear{Ha, Dai, and Le}{Ha et~al\mbox{.}}{2016}]%
        {ha2016hypernetworks}
\bibfield{author}{\bibinfo{person}{David Ha}, \bibinfo{person}{Andrew Dai}, {and} \bibinfo{person}{Quoc~V Le}.} \bibinfo{year}{2016}\natexlab{}.
\newblock \showarticletitle{Hypernetworks}.
\newblock \bibinfo{journal}{\emph{arXiv preprint arXiv:1609.09106}} (\bibinfo{year}{2016}).
\newblock


\bibitem[\protect\citeauthoryear{Han, Zhang, Fu, and Zhou}{Han et~al\mbox{.}}{2022}]%
        {han2022trusted}
\bibfield{author}{\bibinfo{person}{Zongbo Han}, \bibinfo{person}{Changqing Zhang}, \bibinfo{person}{Huazhu Fu}, {and} \bibinfo{person}{Joey~Tianyi Zhou}.} \bibinfo{year}{2022}\natexlab{}.
\newblock \showarticletitle{Trusted multi-view classification with dynamic evidential fusion}.
\newblock \bibinfo{journal}{\emph{IEEE transactions on pattern analysis and machine intelligence}} \bibinfo{volume}{45}, \bibinfo{number}{2} (\bibinfo{year}{2022}), \bibinfo{pages}{2551--2566}.
\newblock


\bibitem[\protect\citeauthoryear{Haroush, Frostig, Heller, and Soudry}{Haroush et~al\mbox{.}}{2021}]%
        {haroush2021statistical}
\bibfield{author}{\bibinfo{person}{Matan Haroush}, \bibinfo{person}{Tzivel Frostig}, \bibinfo{person}{Ruth Heller}, {and} \bibinfo{person}{Daniel Soudry}.} \bibinfo{year}{2021}\natexlab{}.
\newblock \showarticletitle{Statistical testing for efficient out of distribution detection in deep neural networks}.
\newblock \bibinfo{journal}{\emph{arXiv preprint arXiv:2102.12967}} (\bibinfo{year}{2021}).
\newblock


\bibitem[\protect\citeauthoryear{He, Zhang, Ren, and Sun}{He et~al\mbox{.}}{2016}]%
        {he2016deep}
\bibfield{author}{\bibinfo{person}{Kaiming He}, \bibinfo{person}{Xiangyu Zhang}, \bibinfo{person}{Shaoqing Ren}, {and} \bibinfo{person}{Jian Sun}.} \bibinfo{year}{2016}\natexlab{}.
\newblock \showarticletitle{Deep residual learning for image recognition}. In \bibinfo{booktitle}{\emph{Proceedings of the IEEE conference on computer vision and pattern recognition}}. \bibinfo{pages}{770--778}.
\newblock


\bibitem[\protect\citeauthoryear{Hilal, Gadsden, and Yawney}{Hilal et~al\mbox{.}}{2022}]%
        {hilal2022financial}
\bibfield{author}{\bibinfo{person}{Waleed Hilal}, \bibinfo{person}{S~Andrew Gadsden}, {and} \bibinfo{person}{John Yawney}.} \bibinfo{year}{2022}\natexlab{}.
\newblock \showarticletitle{Financial Fraud:: A Review of Anomaly Detection Techniques and Recent Advances}.
\newblock  (\bibinfo{year}{2022}).
\newblock


\bibitem[\protect\citeauthoryear{Huang, Koh, Dobbie, and Pears}{Huang et~al\mbox{.}}{2014}]%
        {huang2014detecting}
\bibfield{author}{\bibinfo{person}{David Tse~Jung Huang}, \bibinfo{person}{Yun~Sing Koh}, \bibinfo{person}{Gillian Dobbie}, {and} \bibinfo{person}{Russel Pears}.} \bibinfo{year}{2014}\natexlab{}.
\newblock \showarticletitle{Detecting volatility shift in data streams}. In \bibinfo{booktitle}{\emph{2014 IEEE International Conference on Data Mining}}. IEEE, \bibinfo{pages}{863--868}.
\newblock


\bibitem[\protect\citeauthoryear{Huang and Kasiviswanathan}{Huang and Kasiviswanathan}{2015}]%
        {huang2015streaming}
\bibfield{author}{\bibinfo{person}{Hao Huang} {and} \bibinfo{person}{Shiva~Prasad Kasiviswanathan}.} \bibinfo{year}{2015}\natexlab{}.
\newblock \showarticletitle{Streaming anomaly detection using randomized matrix sketching}.
\newblock \bibinfo{journal}{\emph{Proceedings of the VLDB Endowment}} \bibinfo{volume}{9}, \bibinfo{number}{3} (\bibinfo{year}{2015}), \bibinfo{pages}{192--203}.
\newblock


\bibitem[\protect\citeauthoryear{Jsang}{Jsang}{2018}]%
        {jsang2018subjective}
\bibfield{author}{\bibinfo{person}{Audun Jsang}.} \bibinfo{year}{2018}\natexlab{}.
\newblock \bibinfo{booktitle}{\emph{Subjective Logic: A formalism for reasoning under uncertainty}}.
\newblock \bibinfo{publisher}{Springer Publishing Company, Incorporated}.
\newblock


\bibitem[\protect\citeauthoryear{Kieu, Yang, Guo, and Jensen}{Kieu et~al\mbox{.}}{2019}]%
        {kieu2019outlier}
\bibfield{author}{\bibinfo{person}{Tung Kieu}, \bibinfo{person}{Bin Yang}, \bibinfo{person}{Chenjuan Guo}, {and} \bibinfo{person}{Christian~S Jensen}.} \bibinfo{year}{2019}\natexlab{}.
\newblock \showarticletitle{Outlier Detection for Time Series with Recurrent Autoencoder Ensembles.}. In \bibinfo{booktitle}{\emph{IJCAI}}. \bibinfo{pages}{2725--2732}.
\newblock


\bibitem[\protect\citeauthoryear{Kim, Shim, Lim, Jeon, Choi, Kim, and Yoon}{Kim et~al\mbox{.}}{2020}]%
        {kim2020rapp}
\bibfield{author}{\bibinfo{person}{Ki~Hyun Kim}, \bibinfo{person}{Sangwoo Shim}, \bibinfo{person}{Yongsub Lim}, \bibinfo{person}{Jongseob Jeon}, \bibinfo{person}{Jeongwoo Choi}, \bibinfo{person}{Byungchan Kim}, {and} \bibinfo{person}{Andre~S Yoon}.} \bibinfo{year}{2020}\natexlab{}.
\newblock \showarticletitle{Rapp: Novelty detection with reconstruction along projection pathway}. In \bibinfo{booktitle}{\emph{International Conference on Learning Representations}}.
\newblock


\bibitem[\protect\citeauthoryear{Kingma and Ba}{Kingma and Ba}{2015}]%
        {adam}
\bibfield{author}{\bibinfo{person}{Diederik~P. Kingma} {and} \bibinfo{person}{Jimmy Ba}.} \bibinfo{year}{2015}\natexlab{}.
\newblock \showarticletitle{Adam: {A} Method for Stochastic Optimization}. In \bibinfo{booktitle}{\emph{3rd International Conference on Learning Representations, {ICLR}}}.
\newblock


\bibitem[\protect\citeauthoryear{Kingma and Welling}{Kingma and Welling}{2013}]%
        {kingma2013auto}
\bibfield{author}{\bibinfo{person}{Diederik~P Kingma} {and} \bibinfo{person}{Max Welling}.} \bibinfo{year}{2013}\natexlab{}.
\newblock \showarticletitle{Auto-encoding variational bayes}.
\newblock \bibinfo{journal}{\emph{arXiv preprint arXiv:1312.6114}} (\bibinfo{year}{2013}).
\newblock


\bibitem[\protect\citeauthoryear{Klinker}{Klinker}{2011}]%
        {klinker2011exponential}
\bibfield{author}{\bibinfo{person}{Frank Klinker}.} \bibinfo{year}{2011}\natexlab{}.
\newblock \showarticletitle{Exponential moving average versus moving exponential average}.
\newblock \bibinfo{journal}{\emph{Mathematische Semesterberichte}}  \bibinfo{volume}{58} (\bibinfo{year}{2011}), \bibinfo{pages}{97--107}.
\newblock


\bibitem[\protect\citeauthoryear{Kloft and Laskov}{Kloft and Laskov}{2010}]%
        {kloft2010online}
\bibfield{author}{\bibinfo{person}{Marius Kloft} {and} \bibinfo{person}{Pavel Laskov}.} \bibinfo{year}{2010}\natexlab{}.
\newblock \showarticletitle{Online anomaly detection under adversarial impact}. In \bibinfo{booktitle}{\emph{Proceedings of the thirteenth international conference on artificial intelligence and statistics}}. JMLR Workshop and Conference Proceedings, \bibinfo{pages}{405--412}.
\newblock


\bibitem[\protect\citeauthoryear{Kloft and Laskov}{Kloft and Laskov}{2012}]%
        {kloft2012security}
\bibfield{author}{\bibinfo{person}{Marius Kloft} {and} \bibinfo{person}{Pavel Laskov}.} \bibinfo{year}{2012}\natexlab{}.
\newblock \showarticletitle{Security analysis of online centroid anomaly detection}.
\newblock \bibinfo{journal}{\emph{The Journal of Machine Learning Research}} \bibinfo{volume}{13}, \bibinfo{number}{1} (\bibinfo{year}{2012}), \bibinfo{pages}{3681--3724}.
\newblock


\bibitem[\protect\citeauthoryear{Lai, Zou, and Lerman}{Lai et~al\mbox{.}}{2019}]%
        {lai2019robust}
\bibfield{author}{\bibinfo{person}{Chieh-Hsin Lai}, \bibinfo{person}{Dongmian Zou}, {and} \bibinfo{person}{Gilad Lerman}.} \bibinfo{year}{2019}\natexlab{}.
\newblock \showarticletitle{Robust subspace recovery layer for unsupervised anomaly detection}.
\newblock \bibinfo{journal}{\emph{arXiv preprint arXiv:1904.00152}} (\bibinfo{year}{2019}).
\newblock


\bibitem[\protect\citeauthoryear{LeCun, Bottou, Bengio, and Haffner}{LeCun et~al\mbox{.}}{1998}]%
        {lecun1998gradient}
\bibfield{author}{\bibinfo{person}{Yann LeCun}, \bibinfo{person}{L{\'e}on Bottou}, \bibinfo{person}{Yoshua Bengio}, {and} \bibinfo{person}{Patrick Haffner}.} \bibinfo{year}{1998}\natexlab{}.
\newblock \showarticletitle{Gradient-based learning applied to document recognition}.
\newblock \bibinfo{journal}{\emph{Proc. IEEE}} \bibinfo{volume}{86}, \bibinfo{number}{11} (\bibinfo{year}{1998}), \bibinfo{pages}{2278--2324}.
\newblock


\bibitem[\protect\citeauthoryear{Li, Yan, Wang, and Jin}{Li et~al\mbox{.}}{2020}]%
        {li2020anomaly}
\bibfield{author}{\bibinfo{person}{Longyuan Li}, \bibinfo{person}{Junchi Yan}, \bibinfo{person}{Haiyang Wang}, {and} \bibinfo{person}{Yaohui Jin}.} \bibinfo{year}{2020}\natexlab{}.
\newblock \showarticletitle{Anomaly detection of time series with smoothness-inducing sequential variational auto-encoder}.
\newblock \bibinfo{journal}{\emph{IEEE transactions on neural networks and learning systems}} \bibinfo{volume}{32}, \bibinfo{number}{3} (\bibinfo{year}{2020}), \bibinfo{pages}{1177--1191}.
\newblock


\bibitem[\protect\citeauthoryear{Li, Yin, Li, Li, Liu, and Zhu}{Li et~al\mbox{.}}{2022}]%
        {li2022unsupervised}
\bibfield{author}{\bibinfo{person}{Sainan Li}, \bibinfo{person}{Qilei Yin}, \bibinfo{person}{Guoliang Li}, \bibinfo{person}{Qi Li}, \bibinfo{person}{Zhuotao Liu}, {and} \bibinfo{person}{Jinwei Zhu}.} \bibinfo{year}{2022}\natexlab{}.
\newblock \showarticletitle{Unsupervised Contextual Anomaly Detection for Database Systems}. In \bibinfo{booktitle}{\emph{Proceedings of the 2022 International Conference on Management of Data}}. \bibinfo{pages}{788--802}.
\newblock


\bibitem[\protect\citeauthoryear{Liu, Ting, and Zhou}{Liu et~al\mbox{.}}{2008}]%
        {liu2008isolation}
\bibfield{author}{\bibinfo{person}{Fei~Tony Liu}, \bibinfo{person}{Kai~Ming Ting}, {and} \bibinfo{person}{Zhi-Hua Zhou}.} \bibinfo{year}{2008}\natexlab{}.
\newblock \showarticletitle{Isolation forest}. In \bibinfo{booktitle}{\emph{2008 eighth ieee international conference on data mining}}. IEEE, \bibinfo{pages}{413--422}.
\newblock


\bibitem[\protect\citeauthoryear{Lu, Liu, Dong, Gu, Gama, and Zhang}{Lu et~al\mbox{.}}{2018}]%
        {lu2018learning}
\bibfield{author}{\bibinfo{person}{Jie Lu}, \bibinfo{person}{Anjin Liu}, \bibinfo{person}{Fan Dong}, \bibinfo{person}{Feng Gu}, \bibinfo{person}{Joao Gama}, {and} \bibinfo{person}{Guangquan Zhang}.} \bibinfo{year}{2018}\natexlab{}.
\newblock \showarticletitle{Learning under concept drift: A review}.
\newblock \bibinfo{journal}{\emph{IEEE transactions on knowledge and data engineering}} \bibinfo{volume}{31}, \bibinfo{number}{12} (\bibinfo{year}{2018}), \bibinfo{pages}{2346--2363}.
\newblock


\bibitem[\protect\citeauthoryear{Lu, Wu, Mueen, Zuluaga, and Keogh}{Lu et~al\mbox{.}}{2022}]%
        {lu2022matrix}
\bibfield{author}{\bibinfo{person}{Yue Lu}, \bibinfo{person}{Renjie Wu}, \bibinfo{person}{Abdullah Mueen}, \bibinfo{person}{Maria~A Zuluaga}, {and} \bibinfo{person}{Eamonn Keogh}.} \bibinfo{year}{2022}\natexlab{}.
\newblock \showarticletitle{Matrix profile XXIV: scaling time series anomaly detection to trillions of datapoints and ultra-fast arriving data streams}. In \bibinfo{booktitle}{\emph{Proceedings of the 28th ACM SIGKDD Conference on Knowledge Discovery and Data Mining}}. \bibinfo{pages}{1173--1182}.
\newblock


\bibitem[\protect\citeauthoryear{Makhzani and Frey}{Makhzani and Frey}{2015}]%
        {makhzani2015winner}
\bibfield{author}{\bibinfo{person}{Alireza Makhzani} {and} \bibinfo{person}{Brendan~J Frey}.} \bibinfo{year}{2015}\natexlab{}.
\newblock \showarticletitle{Winner-take-all autoencoders}.
\newblock \bibinfo{journal}{\emph{Advances in neural information processing systems}}  \bibinfo{volume}{28} (\bibinfo{year}{2015}).
\newblock


\bibitem[\protect\citeauthoryear{Manzoor, Lamba, and Akoglu}{Manzoor et~al\mbox{.}}{2018}]%
        {manzoor2018xstream}
\bibfield{author}{\bibinfo{person}{Emaad Manzoor}, \bibinfo{person}{Hemank Lamba}, {and} \bibinfo{person}{Leman Akoglu}.} \bibinfo{year}{2018}\natexlab{}.
\newblock \showarticletitle{xstream: Outlier detection in feature-evolving data streams}. In \bibinfo{booktitle}{\emph{Proceedings of the 24th ACM SIGKDD International Conference on Knowledge Discovery \& Data Mining}}. \bibinfo{pages}{1963--1972}.
\newblock


\bibitem[\protect\citeauthoryear{Mirsky, Doitshman, Elovici, and Shabtai}{Mirsky et~al\mbox{.}}{2018}]%
        {mirsky2018kitsune}
\bibfield{author}{\bibinfo{person}{Yisroel Mirsky}, \bibinfo{person}{Tomer Doitshman}, \bibinfo{person}{Yuval Elovici}, {and} \bibinfo{person}{Asaf Shabtai}.} \bibinfo{year}{2018}\natexlab{}.
\newblock \showarticletitle{Kitsune: an ensemble of autoencoders for online network intrusion detection}.
\newblock \bibinfo{journal}{\emph{arXiv preprint arXiv:1802.09089}} (\bibinfo{year}{2018}).
\newblock


\bibitem[\protect\citeauthoryear{Na, Kim, and Yu}{Na et~al\mbox{.}}{2018}]%
        {na2018dilof}
\bibfield{author}{\bibinfo{person}{Gyoung~S Na}, \bibinfo{person}{Donghyun Kim}, {and} \bibinfo{person}{Hwanjo Yu}.} \bibinfo{year}{2018}\natexlab{}.
\newblock \showarticletitle{Dilof: Effective and memory efficient local outlier detection in data streams}. In \bibinfo{booktitle}{\emph{Proceedings of the 24th ACM SIGKDD International Conference on Knowledge Discovery \& Data Mining}}. \bibinfo{pages}{1993--2002}.
\newblock


\bibitem[\protect\citeauthoryear{Nawir, Amir, Yaakob, and Lynn}{Nawir et~al\mbox{.}}{2019}]%
        {nawir2019effective}
\bibfield{author}{\bibinfo{person}{Mukrimah Nawir}, \bibinfo{person}{Amiza Amir}, \bibinfo{person}{Naimah Yaakob}, {and} \bibinfo{person}{Ong~Bi Lynn}.} \bibinfo{year}{2019}\natexlab{}.
\newblock \showarticletitle{Effective and efficient network anomaly detection system using machine learning algorithm}.
\newblock \bibinfo{journal}{\emph{Bulletin of Electrical Engineering and Informatics}} \bibinfo{volume}{8}, \bibinfo{number}{1} (\bibinfo{year}{2019}), \bibinfo{pages}{46--51}.
\newblock


\bibitem[\protect\citeauthoryear{Ng, Tian, and Tang}{Ng et~al\mbox{.}}{2011}]%
        {ng2011dirichlet}
\bibfield{author}{\bibinfo{person}{Kai~Wang Ng}, \bibinfo{person}{Guo-Liang Tian}, {and} \bibinfo{person}{Man-Lai Tang}.} \bibinfo{year}{2011}\natexlab{}.
\newblock \showarticletitle{Dirichlet and related distributions: Theory, methods and applications}.
\newblock  (\bibinfo{year}{2011}).
\newblock


\bibitem[\protect\citeauthoryear{Oliner and Stearley}{Oliner and Stearley}{2007}]%
        {4273008}
\bibfield{author}{\bibinfo{person}{Adam Oliner} {and} \bibinfo{person}{Jon Stearley}.} \bibinfo{year}{2007}\natexlab{}.
\newblock \showarticletitle{What Supercomputers Say: A Study of Five System Logs}. In \bibinfo{booktitle}{\emph{37th Annual IEEE/IFIP International Conference on Dependable Systems and Networks (DSN'07)}}. \bibinfo{pages}{575--584}.
\newblock
\urldef\tempurl%
\url{https://doi.org/10.1109/DSN.2007.103}
\showDOI{\tempurl}


\bibitem[\protect\citeauthoryear{Pang, Shen, Cao, and Hengel}{Pang et~al\mbox{.}}{2021}]%
        {pang2021deep}
\bibfield{author}{\bibinfo{person}{Guansong Pang}, \bibinfo{person}{Chunhua Shen}, \bibinfo{person}{Longbing Cao}, {and} \bibinfo{person}{Anton Van~Den Hengel}.} \bibinfo{year}{2021}\natexlab{}.
\newblock \showarticletitle{Deep learning for anomaly detection: A review}.
\newblock \bibinfo{journal}{\emph{ACM computing surveys (CSUR)}} \bibinfo{volume}{54}, \bibinfo{number}{2} (\bibinfo{year}{2021}), \bibinfo{pages}{1--38}.
\newblock


\bibitem[\protect\citeauthoryear{Paparrizos, Kang, Boniol, Tsay, Palpanas, and Franklin}{Paparrizos et~al\mbox{.}}{2022}]%
        {paparrizos2022tsb}
\bibfield{author}{\bibinfo{person}{John Paparrizos}, \bibinfo{person}{Yuhao Kang}, \bibinfo{person}{Paul Boniol}, \bibinfo{person}{Ruey~S Tsay}, \bibinfo{person}{Themis Palpanas}, {and} \bibinfo{person}{Michael~J Franklin}.} \bibinfo{year}{2022}\natexlab{}.
\newblock \showarticletitle{TSB-UAD: an end-to-end benchmark suite for univariate time-series anomaly detection}.
\newblock \bibinfo{journal}{\emph{Proceedings of the VLDB Endowment}} \bibinfo{volume}{15}, \bibinfo{number}{8} (\bibinfo{year}{2022}), \bibinfo{pages}{1697--1711}.
\newblock


\bibitem[\protect\citeauthoryear{Pedregosa, Varoquaux, Gramfort, Michel, Thirion, Grisel, Blondel, Prettenhofer, Weiss, Dubourg, et~al\mbox{.}}{Pedregosa et~al\mbox{.}}{2011}]%
        {pedregosa2011scikit}
\bibfield{author}{\bibinfo{person}{Fabian Pedregosa}, \bibinfo{person}{Ga{\"e}l Varoquaux}, \bibinfo{person}{Alexandre Gramfort}, \bibinfo{person}{Vincent Michel}, \bibinfo{person}{Bertrand Thirion}, \bibinfo{person}{Olivier Grisel}, \bibinfo{person}{Mathieu Blondel}, \bibinfo{person}{Peter Prettenhofer}, \bibinfo{person}{Ron Weiss}, \bibinfo{person}{Vincent Dubourg}, {et~al\mbox{.}}} \bibinfo{year}{2011}\natexlab{}.
\newblock \showarticletitle{Scikit-learn: Machine learning in Python}.
\newblock \bibinfo{journal}{\emph{the Journal of machine Learning research}}  \bibinfo{volume}{12} (\bibinfo{year}{2011}), \bibinfo{pages}{2825--2830}.
\newblock


\bibitem[\protect\citeauthoryear{Pevn{\`y}}{Pevn{\`y}}{2016}]%
        {pevny2016loda}
\bibfield{author}{\bibinfo{person}{Tom{\'a}{\v{s}} Pevn{\`y}}.} \bibinfo{year}{2016}\natexlab{}.
\newblock \showarticletitle{Loda: Lightweight on-line detector of anomalies}.
\newblock \bibinfo{journal}{\emph{Machine Learning}}  \bibinfo{volume}{102} (\bibinfo{year}{2016}), \bibinfo{pages}{275--304}.
\newblock


\bibitem[\protect\citeauthoryear{Rayana}{Rayana}{2016}]%
        {Rayana:2016}
\bibfield{author}{\bibinfo{person}{Shebuti Rayana}.} \bibinfo{year}{2016}\natexlab{}.
\newblock \bibinfo{title}{ODDS Library}.
\newblock
\newblock
\newblock
\shownote{\url{https://odds.cs.stonybrook.edu}. Accessed:2023-07.}


\bibitem[\protect\citeauthoryear{Ruff, Kauffmann, Vandermeulen, Montavon, Samek, Kloft, Dietterich, and M{\"u}ller}{Ruff et~al\mbox{.}}{2021}]%
        {ruff2021unifying}
\bibfield{author}{\bibinfo{person}{Lukas Ruff}, \bibinfo{person}{Jacob~R Kauffmann}, \bibinfo{person}{Robert~A Vandermeulen}, \bibinfo{person}{Gr{\'e}goire Montavon}, \bibinfo{person}{Wojciech Samek}, \bibinfo{person}{Marius Kloft}, \bibinfo{person}{Thomas~G Dietterich}, {and} \bibinfo{person}{Klaus-Robert M{\"u}ller}.} \bibinfo{year}{2021}\natexlab{}.
\newblock \showarticletitle{A unifying review of deep and shallow anomaly detection}.
\newblock \bibinfo{journal}{\emph{Proc. IEEE}} \bibinfo{volume}{109}, \bibinfo{number}{5} (\bibinfo{year}{2021}), \bibinfo{pages}{756--795}.
\newblock


\bibitem[\protect\citeauthoryear{{\v{S}}abi{\'c}, Keeley, Henderson, and Nannemann}{{\v{S}}abi{\'c} et~al\mbox{.}}{2021}]%
        {vsabic2021healthcare}
\bibfield{author}{\bibinfo{person}{Edin {\v{S}}abi{\'c}}, \bibinfo{person}{David Keeley}, \bibinfo{person}{Bailey Henderson}, {and} \bibinfo{person}{Sara Nannemann}.} \bibinfo{year}{2021}\natexlab{}.
\newblock \showarticletitle{Healthcare and anomaly detection: using machine learning to predict anomalies in heart rate data}.
\newblock \bibinfo{journal}{\emph{AI \& SOCIETY}} \bibinfo{volume}{36}, \bibinfo{number}{1} (\bibinfo{year}{2021}), \bibinfo{pages}{149--158}.
\newblock


\bibitem[\protect\citeauthoryear{Sabokrou, Fathy, and Hoseini}{Sabokrou et~al\mbox{.}}{2016}]%
        {sabokrou2016video}
\bibfield{author}{\bibinfo{person}{Mohammad Sabokrou}, \bibinfo{person}{Mahmood Fathy}, {and} \bibinfo{person}{Mojtaba Hoseini}.} \bibinfo{year}{2016}\natexlab{}.
\newblock \showarticletitle{Video anomaly detection and localisation based on the sparsity and reconstruction error of auto-encoder}.
\newblock \bibinfo{journal}{\emph{Electronics Letters}} \bibinfo{volume}{52}, \bibinfo{number}{13} (\bibinfo{year}{2016}), \bibinfo{pages}{1122--1124}.
\newblock


\bibitem[\protect\citeauthoryear{Sakurada and Yairi}{Sakurada and Yairi}{2014}]%
        {sakurada2014anomaly}
\bibfield{author}{\bibinfo{person}{Mayu Sakurada} {and} \bibinfo{person}{Takehisa Yairi}.} \bibinfo{year}{2014}\natexlab{}.
\newblock \showarticletitle{Anomaly detection using autoencoders with nonlinear dimensionality reduction}. In \bibinfo{booktitle}{\emph{Proceedings of the MLSDA 2014 2nd workshop on machine learning for sensory data analysis}}. \bibinfo{pages}{4--11}.
\newblock


\bibitem[\protect\citeauthoryear{Salehi, Leckie, Bezdek, Vaithianathan, and Zhang}{Salehi et~al\mbox{.}}{2016}]%
        {salehi2016fast}
\bibfield{author}{\bibinfo{person}{Mahsa Salehi}, \bibinfo{person}{Christopher Leckie}, \bibinfo{person}{James~C Bezdek}, \bibinfo{person}{Tharshan Vaithianathan}, {and} \bibinfo{person}{Xuyun Zhang}.} \bibinfo{year}{2016}\natexlab{}.
\newblock \showarticletitle{Fast memory efficient local outlier detection in data streams}.
\newblock \bibinfo{journal}{\emph{IEEE Transactions on Knowledge and Data Engineering}} \bibinfo{volume}{28}, \bibinfo{number}{12} (\bibinfo{year}{2016}), \bibinfo{pages}{3246--3260}.
\newblock


\bibitem[\protect\citeauthoryear{Sathe and Aggarwal}{Sathe and Aggarwal}{2016}]%
        {sathe2016subspace}
\bibfield{author}{\bibinfo{person}{Saket Sathe} {and} \bibinfo{person}{Charu~C Aggarwal}.} \bibinfo{year}{2016}\natexlab{}.
\newblock \showarticletitle{Subspace outlier detection in linear time with randomized hashing}. In \bibinfo{booktitle}{\emph{2016 IEEE 16th International Conference on Data Mining (ICDM)}}. IEEE, \bibinfo{pages}{459--468}.
\newblock


\bibitem[\protect\citeauthoryear{Savage, Zhang, Yu, Chou, and Wang}{Savage et~al\mbox{.}}{2014}]%
        {savage2014anomaly}
\bibfield{author}{\bibinfo{person}{David Savage}, \bibinfo{person}{Xiuzhen Zhang}, \bibinfo{person}{Xinghuo Yu}, \bibinfo{person}{Pauline Chou}, {and} \bibinfo{person}{Qingmai Wang}.} \bibinfo{year}{2014}\natexlab{}.
\newblock \showarticletitle{Anomaly detection in online social networks}.
\newblock \bibinfo{journal}{\emph{Social networks}}  \bibinfo{volume}{39} (\bibinfo{year}{2014}), \bibinfo{pages}{62--70}.
\newblock


\bibitem[\protect\citeauthoryear{Schmidl, Wenig, and Papenbrock}{Schmidl et~al\mbox{.}}{2022}]%
        {schmidl2022anomaly}
\bibfield{author}{\bibinfo{person}{Sebastian Schmidl}, \bibinfo{person}{Phillip Wenig}, {and} \bibinfo{person}{Thorsten Papenbrock}.} \bibinfo{year}{2022}\natexlab{}.
\newblock \showarticletitle{Anomaly detection in time series: a comprehensive evaluation}.
\newblock \bibinfo{journal}{\emph{Proceedings of the VLDB Endowment}} \bibinfo{volume}{15}, \bibinfo{number}{9} (\bibinfo{year}{2022}), \bibinfo{pages}{1779--1797}.
\newblock


\bibitem[\protect\citeauthoryear{Sensoy, Kaplan, and Kandemir}{Sensoy et~al\mbox{.}}{2018}]%
        {sensoy2018evidential}
\bibfield{author}{\bibinfo{person}{Murat Sensoy}, \bibinfo{person}{Lance Kaplan}, {and} \bibinfo{person}{Melih Kandemir}.} \bibinfo{year}{2018}\natexlab{}.
\newblock \showarticletitle{Evidential deep learning to quantify classification uncertainty}.
\newblock \bibinfo{journal}{\emph{Advances in neural information processing systems}}  \bibinfo{volume}{31} (\bibinfo{year}{2018}).
\newblock


\bibitem[\protect\citeauthoryear{Soleimany, Amini, Goldman, Rus, Bhatia, and Coley}{Soleimany et~al\mbox{.}}{2021}]%
        {soleimany2021evidential}
\bibfield{author}{\bibinfo{person}{Ava~P Soleimany}, \bibinfo{person}{Alexander Amini}, \bibinfo{person}{Samuel Goldman}, \bibinfo{person}{Daniela Rus}, \bibinfo{person}{Sangeeta~N Bhatia}, {and} \bibinfo{person}{Connor~W Coley}.} \bibinfo{year}{2021}\natexlab{}.
\newblock \showarticletitle{Evidential deep learning for guided molecular property prediction and discovery}.
\newblock \bibinfo{journal}{\emph{ACS central science}} \bibinfo{volume}{7}, \bibinfo{number}{8} (\bibinfo{year}{2021}), \bibinfo{pages}{1356--1367}.
\newblock


\bibitem[\protect\citeauthoryear{Souza, dos Reis, Maletzke, and Batista}{Souza et~al\mbox{.}}{2020}]%
        {souza2020challenges}
\bibfield{author}{\bibinfo{person}{Vinicius~MA Souza}, \bibinfo{person}{Denis~M dos Reis}, \bibinfo{person}{Andre~G Maletzke}, {and} \bibinfo{person}{Gustavo~EAPA Batista}.} \bibinfo{year}{2020}\natexlab{}.
\newblock \showarticletitle{Challenges in benchmarking stream learning algorithms with real-world data}.
\newblock \bibinfo{journal}{\emph{Data Mining and Knowledge Discovery}}  \bibinfo{volume}{34} (\bibinfo{year}{2020}), \bibinfo{pages}{1805--1858}.
\newblock


\bibitem[\protect\citeauthoryear{Tan, Ting, and Liu}{Tan et~al\mbox{.}}{2011}]%
        {tan2011fast}
\bibfield{author}{\bibinfo{person}{Swee~Chuan Tan}, \bibinfo{person}{Kai~Ming Ting}, {and} \bibinfo{person}{Tony~Fei Liu}.} \bibinfo{year}{2011}\natexlab{}.
\newblock \showarticletitle{Fast anomaly detection for streaming data}. In \bibinfo{booktitle}{\emph{Twenty-second international joint conference on artificial intelligence}}. Citeseer.
\newblock


\bibitem[\protect\citeauthoryear{Tartakovsky, Polunchenko, and Sokolov}{Tartakovsky et~al\mbox{.}}{2012}]%
        {tartakovsky2012efficient}
\bibfield{author}{\bibinfo{person}{Alexander~G Tartakovsky}, \bibinfo{person}{Aleksey~S Polunchenko}, {and} \bibinfo{person}{Grigory Sokolov}.} \bibinfo{year}{2012}\natexlab{}.
\newblock \showarticletitle{Efficient computer network anomaly detection by changepoint detection methods}.
\newblock \bibinfo{journal}{\emph{IEEE Journal of Selected Topics in Signal Processing}} \bibinfo{volume}{7}, \bibinfo{number}{1} (\bibinfo{year}{2012}), \bibinfo{pages}{4--11}.
\newblock


\bibitem[\protect\citeauthoryear{Tavallaee, Bagheri, Lu, and Ghorbani}{Tavallaee et~al\mbox{.}}{2009}]%
        {tavallaee2009detailed}
\bibfield{author}{\bibinfo{person}{Mahbod Tavallaee}, \bibinfo{person}{Ebrahim Bagheri}, \bibinfo{person}{Wei Lu}, {and} \bibinfo{person}{Ali~A Ghorbani}.} \bibinfo{year}{2009}\natexlab{}.
\newblock \showarticletitle{A detailed analysis of the KDD CUP 99 data set}. In \bibinfo{booktitle}{\emph{2009 IEEE symposium on computational intelligence for security and defense applications}}. Ieee, \bibinfo{pages}{1--6}.
\newblock


\bibitem[\protect\citeauthoryear{Tian, Azarian, and Pecht}{Tian et~al\mbox{.}}{2014}]%
        {tian2014anomaly}
\bibfield{author}{\bibinfo{person}{Jing Tian}, \bibinfo{person}{Michael~H Azarian}, {and} \bibinfo{person}{Michael Pecht}.} \bibinfo{year}{2014}\natexlab{}.
\newblock \showarticletitle{Anomaly detection using self-organizing maps-based k-nearest neighbor algorithm}. In \bibinfo{booktitle}{\emph{PHM society European conference}}, Vol.~\bibinfo{volume}{2}.
\newblock


\bibitem[\protect\citeauthoryear{Toliopoulos, Bellas, Gounaris, and Papadopoulos}{Toliopoulos et~al\mbox{.}}{2020}]%
        {toliopoulos2020proud}
\bibfield{author}{\bibinfo{person}{Theodoros Toliopoulos}, \bibinfo{person}{Christos Bellas}, \bibinfo{person}{Anastasios Gounaris}, {and} \bibinfo{person}{Apostolos Papadopoulos}.} \bibinfo{year}{2020}\natexlab{}.
\newblock \showarticletitle{PROUD: parallel outlier detection for streams}. In \bibinfo{booktitle}{\emph{Proceedings of the 2020 ACM SIGMOD International Conference on Management of Data}}. \bibinfo{pages}{2717--2720}.
\newblock


\bibitem[\protect\citeauthoryear{Tran, Fan, and Shahabi}{Tran et~al\mbox{.}}{2016}]%
        {tran2016distance}
\bibfield{author}{\bibinfo{person}{Luan Tran}, \bibinfo{person}{Liyue Fan}, {and} \bibinfo{person}{Cyrus Shahabi}.} \bibinfo{year}{2016}\natexlab{}.
\newblock \showarticletitle{Distance-based outlier detection in data streams}.
\newblock \bibinfo{journal}{\emph{Proceedings of the VLDB Endowment}} \bibinfo{volume}{9}, \bibinfo{number}{12} (\bibinfo{year}{2016}), \bibinfo{pages}{1089--1100}.
\newblock


\bibitem[\protect\citeauthoryear{Tran, Mun, and Shahabi}{Tran et~al\mbox{.}}{2020}]%
        {tran2020real}
\bibfield{author}{\bibinfo{person}{Luan Tran}, \bibinfo{person}{Min~Y Mun}, {and} \bibinfo{person}{Cyrus Shahabi}.} \bibinfo{year}{2020}\natexlab{}.
\newblock \showarticletitle{Real-time distance-based outlier detection in data streams}.
\newblock \bibinfo{journal}{\emph{Proceedings of the VLDB Endowment}} \bibinfo{volume}{14}, \bibinfo{number}{2} (\bibinfo{year}{2020}), \bibinfo{pages}{141--153}.
\newblock


\bibitem[\protect\citeauthoryear{Vu, Ueta, Hashimoto, Maeno, Pranata, and Shen}{Vu et~al\mbox{.}}{2019}]%
        {vu2019anomaly}
\bibfield{author}{\bibinfo{person}{Ha~Son Vu}, \bibinfo{person}{Daisuke Ueta}, \bibinfo{person}{Kiyoshi Hashimoto}, \bibinfo{person}{Kazuki Maeno}, \bibinfo{person}{Sugiri Pranata}, {and} \bibinfo{person}{Sheng~Mei Shen}.} \bibinfo{year}{2019}\natexlab{}.
\newblock \showarticletitle{Anomaly detection with adversarial dual autoencoders}.
\newblock \bibinfo{journal}{\emph{arXiv preprint arXiv:1902.06924}} (\bibinfo{year}{2019}).
\newblock


\bibitem[\protect\citeauthoryear{Wang, Luo, and Zhou}{Wang et~al\mbox{.}}{2020}]%
        {wang2020guardhealth}
\bibfield{author}{\bibinfo{person}{Ziyu Wang}, \bibinfo{person}{Nanqing Luo}, {and} \bibinfo{person}{Pan Zhou}.} \bibinfo{year}{2020}\natexlab{}.
\newblock \showarticletitle{GuardHealth: Blockchain empowered secure data management and Graph Convolutional Network enabled anomaly detection in smart healthcare}.
\newblock \bibinfo{journal}{\emph{J. Parallel and Distrib. Comput.}}  \bibinfo{volume}{142} (\bibinfo{year}{2020}), \bibinfo{pages}{1--12}.
\newblock


\bibitem[\protect\citeauthoryear{Wu, Shen, and Van Den~Hengel}{Wu et~al\mbox{.}}{2019}]%
        {wu2019wider}
\bibfield{author}{\bibinfo{person}{Zifeng Wu}, \bibinfo{person}{Chunhua Shen}, {and} \bibinfo{person}{Anton Van Den~Hengel}.} \bibinfo{year}{2019}\natexlab{}.
\newblock \showarticletitle{Wider or deeper: Revisiting the resnet model for visual recognition}.
\newblock \bibinfo{journal}{\emph{Pattern Recognition}}  \bibinfo{volume}{90} (\bibinfo{year}{2019}), \bibinfo{pages}{119--133}.
\newblock


\bibitem[\protect\citeauthoryear{Xiao, Rasul, and Vollgraf}{Xiao et~al\mbox{.}}{2017}]%
        {xiao2017fashion}
\bibfield{author}{\bibinfo{person}{Han Xiao}, \bibinfo{person}{Kashif Rasul}, {and} \bibinfo{person}{Roland Vollgraf}.} \bibinfo{year}{2017}\natexlab{}.
\newblock \showarticletitle{Fashion-mnist: a novel image dataset for benchmarking machine learning algorithms}.
\newblock \bibinfo{journal}{\emph{arXiv preprint arXiv:1708.07747}} (\bibinfo{year}{2017}).
\newblock


\bibitem[\protect\citeauthoryear{Yilmaz and Kozat}{Yilmaz and Kozat}{2020}]%
        {yilmaz2020pysad}
\bibfield{author}{\bibinfo{person}{Selim~F Yilmaz} {and} \bibinfo{person}{Suleyman~S Kozat}.} \bibinfo{year}{2020}\natexlab{}.
\newblock \showarticletitle{Pysad: A streaming anomaly detection framework in python}.
\newblock \bibinfo{journal}{\emph{arXiv preprint arXiv:2009.02572}} (\bibinfo{year}{2020}).
\newblock


\bibitem[\protect\citeauthoryear{Yoon, Lee, and Lee}{Yoon et~al\mbox{.}}{2020}]%
        {yoon2020ultrafast}
\bibfield{author}{\bibinfo{person}{Susik Yoon}, \bibinfo{person}{Jae-Gil Lee}, {and} \bibinfo{person}{Byung~Suk Lee}.} \bibinfo{year}{2020}\natexlab{}.
\newblock \showarticletitle{Ultrafast local outlier detection from a data stream with stationary region skipping}. In \bibinfo{booktitle}{\emph{Proceedings of the 26th ACM SIGKDD International Conference on Knowledge Discovery \& Data Mining}}. \bibinfo{pages}{1181--1191}.
\newblock


\bibitem[\protect\citeauthoryear{Yoon, Lee, Lee, and Lee}{Yoon et~al\mbox{.}}{2022}]%
        {yoon2022adaptive}
\bibfield{author}{\bibinfo{person}{Susik Yoon}, \bibinfo{person}{Youngjun Lee}, \bibinfo{person}{Jae-Gil Lee}, {and} \bibinfo{person}{Byung~Suk Lee}.} \bibinfo{year}{2022}\natexlab{}.
\newblock \showarticletitle{Adaptive Model Pooling for Online Deep Anomaly Detection from a Complex Evolving Data Stream}. In \bibinfo{booktitle}{\emph{Proceedings of the 28th ACM SIGKDD Conference on Knowledge Discovery and Data Mining}}. \bibinfo{pages}{2347--2357}.
\newblock


\bibitem[\protect\citeauthoryear{Yoon, Shin, Lee, and Lee}{Yoon et~al\mbox{.}}{2021}]%
        {yoon2021multiple}
\bibfield{author}{\bibinfo{person}{Susik Yoon}, \bibinfo{person}{Yooju Shin}, \bibinfo{person}{Jae-Gil Lee}, {and} \bibinfo{person}{Byung~Suk Lee}.} \bibinfo{year}{2021}\natexlab{}.
\newblock \showarticletitle{Multiple dynamic outlier-detection from a data stream by exploiting duality of data and queries}. In \bibinfo{booktitle}{\emph{Proceedings of the 2021 International Conference on Management of Data}}. \bibinfo{pages}{2063--2075}.
\newblock


\bibitem[\protect\citeauthoryear{Zenati, Romain, Foo, Lecouat, and Chandrasekhar}{Zenati et~al\mbox{.}}{2018}]%
        {zenati2018adversarially}
\bibfield{author}{\bibinfo{person}{Houssam Zenati}, \bibinfo{person}{Manon Romain}, \bibinfo{person}{Chuan-Sheng Foo}, \bibinfo{person}{Bruno Lecouat}, {and} \bibinfo{person}{Vijay Chandrasekhar}.} \bibinfo{year}{2018}\natexlab{}.
\newblock \showarticletitle{Adversarially learned anomaly detection}. In \bibinfo{booktitle}{\emph{2018 IEEE International conference on data mining (ICDM)}}. IEEE, \bibinfo{pages}{727--736}.
\newblock


\bibitem[\protect\citeauthoryear{Zhao, Chen, Yu, Wang, Li, Qiu, Xu, Zhang, Sui, and Pei}{Zhao et~al\mbox{.}}{2021a}]%
        {zhao2021identifying}
\bibfield{author}{\bibinfo{person}{Nengwen Zhao}, \bibinfo{person}{Junjie Chen}, \bibinfo{person}{Zhaoyang Yu}, \bibinfo{person}{Honglin Wang}, \bibinfo{person}{Jiesong Li}, \bibinfo{person}{Bin Qiu}, \bibinfo{person}{Hongyu Xu}, \bibinfo{person}{Wenchi Zhang}, \bibinfo{person}{Kaixin Sui}, {and} \bibinfo{person}{Dan Pei}.} \bibinfo{year}{2021}\natexlab{a}.
\newblock \showarticletitle{Identifying bad software changes via multimodal anomaly detection for online service systems}. In \bibinfo{booktitle}{\emph{Proceedings of the 29th ACM Joint Meeting on European Software Engineering Conference and Symposium on the Foundations of Software Engineering}}. \bibinfo{pages}{527--539}.
\newblock


\bibitem[\protect\citeauthoryear{Zhao, Wang, Li, Peng, Wang, Pan, Wu, Feng, Wen, Zhang, et~al\mbox{.}}{Zhao et~al\mbox{.}}{2021b}]%
        {zhao2021empirical}
\bibfield{author}{\bibinfo{person}{Nengwen Zhao}, \bibinfo{person}{Honglin Wang}, \bibinfo{person}{Zeyan Li}, \bibinfo{person}{Xiao Peng}, \bibinfo{person}{Gang Wang}, \bibinfo{person}{Zhu Pan}, \bibinfo{person}{Yong Wu}, \bibinfo{person}{Zhen Feng}, \bibinfo{person}{Xidao Wen}, \bibinfo{person}{Wenchi Zhang}, {et~al\mbox{.}}} \bibinfo{year}{2021}\natexlab{b}.
\newblock \showarticletitle{An empirical investigation of practical log anomaly detection for online service systems}. In \bibinfo{booktitle}{\emph{Proceedings of the 29th ACM Joint Meeting on European Software Engineering Conference and Symposium on the Foundations of Software Engineering}}. \bibinfo{pages}{1404--1415}.
\newblock


\bibitem[\protect\citeauthoryear{Zhao, Nasrullah, and Li}{Zhao et~al\mbox{.}}{2019}]%
        {zhao2019pyod}
\bibfield{author}{\bibinfo{person}{Yue Zhao}, \bibinfo{person}{Zain Nasrullah}, {and} \bibinfo{person}{Zheng Li}.} \bibinfo{year}{2019}\natexlab{}.
\newblock \showarticletitle{Pyod: A python toolbox for scalable outlier detection}.
\newblock \bibinfo{journal}{\emph{arXiv preprint arXiv:1901.01588}} (\bibinfo{year}{2019}).
\newblock


\bibitem[\protect\citeauthoryear{Zheng, Cai, Chua, Wang, Ngiam, and Ooi}{Zheng et~al\mbox{.}}{2020}]%
        {zheng2020tracer}
\bibfield{author}{\bibinfo{person}{Kaiping Zheng}, \bibinfo{person}{Shaofeng Cai}, \bibinfo{person}{Horng~Ruey Chua}, \bibinfo{person}{Wei Wang}, \bibinfo{person}{Kee~Yuan Ngiam}, {and} \bibinfo{person}{Beng~Chin Ooi}.} \bibinfo{year}{2020}\natexlab{}.
\newblock \showarticletitle{Tracer: A framework for facilitating accurate and interpretable analytics for high stakes applications}. In \bibinfo{booktitle}{\emph{Proceedings of the 2020 ACM SIGMOD International Conference on Management of Data}}. \bibinfo{pages}{1747--1763}.
\newblock


\bibitem[\protect\citeauthoryear{Zong, Song, Min, Cheng, Lumezanu, Cho, and Chen}{Zong et~al\mbox{.}}{2018}]%
        {zong2018deep}
\bibfield{author}{\bibinfo{person}{Bo Zong}, \bibinfo{person}{Qi Song}, \bibinfo{person}{Martin~Renqiang Min}, \bibinfo{person}{Wei Cheng}, \bibinfo{person}{Cristian Lumezanu}, \bibinfo{person}{Daeki Cho}, {and} \bibinfo{person}{Haifeng Chen}.} \bibinfo{year}{2018}\natexlab{}.
\newblock \showarticletitle{Deep autoencoding gaussian mixture model for unsupervised anomaly detection}. In \bibinfo{booktitle}{\emph{International conference on learning representations}}.
\newblock


\end{thebibliography}

\end{document}